# From Digital Humanities to Quantum Humanities: Potentials and Applications

**Johanna Barzen**

University of Stuttgart
Institute of Architecture of Application Systems (IAAS)
Stuttgart, Germany

e-mail: johanna.barzen@iaas.uni-stuttgart.de
telephone: + 49 (0)711 685 88487

**Abstract**:
Quantum computers are becoming real. Therefore, it is promising to use their potentials in different applications areas, which includes research in the humanities. Due to an increasing amount of data that needs to be processed in the digital humanities the use of quantum computers can contribute to this research area. To give an impression on how beneficial such involvement of quantum computers can be when analyzing data from the humanities, a use case from the media science is presented. Therefore, both the theoretical basis and the tooling support for analyzing the data from our digital humanities project MUSE is described. This includes a data analysis pipeline, containing e.g. various approaches for data preparation, feature engineering, clustering, and classification where several steps can be realized classically, but also supported by quantum computers.

## 1 Introduction

As quantum computers are becoming real there are several application areas where the use of quantum computers is expected to be especially promising (National Academies of Sciences, Engineering, and Medicine 2019): one is simulation, focusing on molecular properties used in material science or in pharma industry, and the other is machine learning e.g. classification, clustering or dimension reduction making heavy use of optimization.

The main difference between classical computers and quantum computers is the different information unit used, namely bits for classical computers and *qubits* for quantum computers (Nielsen and Chuang 2010). Qubits are not restricted to the two states 0 and 1 but can be in an infinite number of different combinations of these states, referred to as *superposition*. Furthermore, individual qubits can be combined into a quantum register resulting in exponentially growing number of data that such a register can hold. As an example, a quantum register with 50 qubits corresponds to $2^{50}$ possible combinations of the states of the individual qubits which are again in superposition corresponding to approximately a petabyte of classical data. All these values can be manipulated by a single operation all at once, which is called *quantum parallelism*.

Operations on single qubits or quantum registers, as a combination of qubits, are unitary transformations. They can be used to process the data to achieve everything classical computers can perform (Rieffel and Polak 2011) but to also use quantum inherent phenomena, e.g. *entanglement*, to – among other potentials – solve a variety of problems that were practically not solvable before, some even with exponential speedup (Nielsen and Chuang 2002, Preskill 2018). There are different quantum computing models, e.g., gate-based (Michielsen et al. 2017), measurement-based (Jozsa 2006), and adiabatic quantum computing (Aharonov et al. 2008), which represent quantum algorithms in various ways. As it can be shown that the different models are formally equivalent (Aharonov et al. 2008, Jozsa 2006) this chapter restricts the considerations and implementations to the gate-based quantum computing model. Also, quantum computers based on the gate-based quantum computing model are universal quantum computers and most commercially available quantum computers are based on this model (LaRose 2019).

As different vendors, such as IBM or Rigetti, developed quantum computers in recent years, and offer access to them via the cloud (LaRose 2019, Leymann et al. 2020) quantum computing is becoming a lively research field and first real word applications are developed in industry. When taking a closer look at the research areas quantum computing is applied to, there are mainly applications in the natural sciences that can be found, for example general approaches like (Bhaskar et al. 2015), presenting quantum algorithms and circuits for modules implementing fundamental functions (e.g. the square root, the natural logarithm, and arbitrary fractional powers) or more specific approaches as in molecular simulation in the material sciences (Kandala et al. 2017, McClean et al. 2017) or in artificial intelligence and machine learning (Dunjko et al. 2016, Havlicek et al. 2018, Otterbach et al. 2017). Nevertheless, first applications can be identified in the humanities too, e.g. *quantum social science* as interdisciplinary field which draws parallels between quantum physics and the social sciences (Haven and Khrennikov 2013), but this is still to be called rudimentary in respect to the potential quantum computers can have in different areas of the humanities.

The chapter is structured as follows: Section 2 presents the vision for quantum humanities. Therefore, the potential benefits that can be expected but also the current challenges are outlined. Section 3 interduces the digital humanities project MUSE as use case for quantum humanities. As the use case strongly relies on analyzing data section 4 focuses on introducing our data analysis pipeline, how to cope with categorical data, and how pattern languages based on data analysis can be detected. A core concept of data analysis are artificial neural networks. Therefore, section 5 provides a mathematical definition of neurons, neuronal networks, and perceptrons, outlines how they are used for restricted Boltzmann machines and autoencoders, and sketches their realizations on quantum computers. As many quantum algorithms are hybrid, meaning they consist of both, a quantum part and a classical part, section 6 focuses on variational hybrid quantum-classical algorithms. The main idea is outlined as well as an application for clustering, namely a quantum version for solving the maximum cut problem. Therefore, techniques as quantum approximate optimization algorithm (QAOA) and variational quantum eigensolver (VQE) are discussed. To support access to the described techniques, section 7 focuses on our quantum humanities analysis tool (QHAna), that is used for pattern detection. In addition to the substantive goal of the use case from the MUSE project to extract costume patterns in films, QHAna provides easy access for people without quantum computing background to evaluate the benefits of using quantum algorithms and allows them to gain initial application knowledge in the field of quantum humanities. Since both, classical analysis algorithms as well as their quantum counterparts are included, the comparison of their results allows a deeper understanding within quantum humanities. Section 8 concludes the chapter and gives an outlook on future work in quantum humanities.

How to read this chapter: As quantum humanities brings together very heterogenous disciplines (e.g. the respective humanities discipline, physics, mathematics, computer science) and each has its own approaches, terminology, concepts and research culture, a fundamental challenge of quantum humanities is to find a common language to communicate between these disciplines. Thus, the chapter starts with an easy-to-understand introduction to the overall topic, the vision and potentials that does not require any previous knowledge (section 1 - 4). The underlying ideas and concepts get more and more refined, up to mathematical definitions (section 5-6). We use mathematical definitions as they give little room for interpretation and therefore seems appropriate as long as there is no common language yet. As the mathematical definitions may be challenging to some readers they are accomplished by descriptions and pictures giving a high-level view on the core concepts, so some readers may skip over the mathematical formalisms. Please note, that algorithmic details themselves are not the focus. Instead, the underlying concepts of the (classical and quantum) algorithms are provided that are relevant for our approach, as well as the tooling provided by QHAna. QHAna than outlines the practical usage of the introduced concepts in a use case centric manner (section 7).

## 2 Towards Quantum Humanities

How beneficial the use of computers, their practical support as well as techniques and methods from computer science, can contribute to research in the humanities has been proven by the establishment of the digital humanities. With quantum computers becoming available, it is promising to extend the usage of classical computers as done in the digital humanities by the use of quantum computers. As stressed in the introduction, working with quantum computers, programing algorithms and letting them run on quantum hardware is very different to working with classical computers. Therefore, we coined the term *quantum humanities* to describe addressing research in the humanities with methods and techniques from quantum computing (Barzen and Leymann 2019, QH 2021). As the basis for quantum humanities the following section will outline the core components defining digital humanities.

### 2.1 Digital Humanities

When speaking about *digital humanities* the term combines a broad variety of topics, contents and contexts. This is reflected by the continuously growing amount of literature on divergent approaches in this domain (Berry 2012, Terras et al. 2013, Jannidis et al. 2017, Berry and Fagerjord 2017, Brown 2020). But there are three core elements that can be identified when taking a closer look at approaches that define digital humanities: (i) digital humanities is about bringing together humanities research and information technology (ii) it's a diverse field and (iii) it's a still emerging field (DHQ 2021).

The first of those highlighted three core elements stresses the combination of computer science and the humanities research. Digital humanities therefore are a highly inter- and transdisciplinary field operating between the traditional humanities and the (in comparison) rather young computer science. The aim is to examine existing and new questions from the humanities by using concepts, methods and tools from computer science. Therefore, digital humanities is understood as a field of methods that unites different topics, approaches and sub-disciplines and as such fosters synergies between different scientific disciplines.

This is rather complex and hints to core element two: digital humanities is a diverse field. 'Humanities' already encompasses a large variety of disciplines, such as philosophy, linguistics, arts, cultural studies, archaeology, musicology and media science, to name just a few. It gets even broader when taking into account that also "computer science" covers quite different subject areas like databases, visualization, or theoretical computer science. In the compilation of so many fields of interest, digital humanities is a very heterogeneous area. It attempts to include the application as well as the further development of tools and methods of computer science in order to possibly obtain new and profound answers to research questions in the humanities while performing a critical self-reflection of the methodological and theoretical approaches.

Even though discussions on digital humanities are around for about 60 years and core areas such as computational linguistics have established themselves through their own courses of study, digital humanities is still a young field of research that has only recently become an independent discipline (Rehbein and Sahle 2013). This is the third core element of many definitions of digital humanities: Digital humanities is a young field whose own definition is constantly being developed (What is Digital Humanities? 2021).

Nevertheless, a broad variety of research programs and projects, conferences and competence centers, stress the benefits of using computers as tools as well as methods from computer science to support research in the humanities. Techniques like databases, visualization, or distant reading have shown how the use of the classical computer can contribute to access knowledge and to broaden the approaches for addressing already stated problems. Now, with quantum computers there is a new type of computer just becoming reality and it allows to further improve the way in which computers can support research in the humanities. Combining the power quantum computer promises with research done in the humanities – based on the achievements in digital humanities research – is what we call *quantum humanities* (Barzen and Leymann 2019).

## 2.2 Potential Benefits of Quantum Humanities

As quantum computers are superior to classical computer in various areas, their use has a high potential contributing to research in the humanities. Especially, since quantum computers are becoming more easily accessible – for example via the cloud (Leymann et al. 2020) – their promising potentials are more and more applicable for first 'proof of concept' application as outlined in (Barzen and Leymann 2020) and will be detailed in section 7. In the following, seven potentials (Barzen and Leymann 2019) will be outlined, which promise to be particularly interesting for supporting digital humanities:

(i): The core benefit of quantum computers is an often stressed potential speedup in several problem areas. Meaning, quantum computers solve certain types of problems much faster (Rønnow et al. 2014) than classical computers (e.g. in the decoding of passwords (Shor 1995), the determination of global properties of Boolean functions (Deutsch 1985) or unstructured search (Grover 1996)). As the amount of data to be processed in the digital humanities is continuously growing, the rapid evaluation of data is becoming increasingly important. Since many algorithms developed for quantum computers can make certain statements about global properties of functions much faster than algorithms developed for classical computers, an enormous time saving can be assumed here. In this context, time saving does not only mean a mere gain in speed of the proper algorithms, but methods of dimensional reduction, for example, that reduce the amount of data to be processed by the proper algorithms, can play a major role here – and the latter can also be performed much more efficiently by quantum computers (e.g. by quantum principal component analysis).

(ii): Quantum computers can process large amounts of data in a single step (Nielsen and Chuang 2010). Due to the superposition of all possible states of the quantum mechanical system, the quantum computer allows true parallelism in the calculation (manipulations), which leads to a significantly higher computing power. This in turn means that complex computational problems, such as those increasingly found in the digital humanities, can be dealt with much more effectively.

(iii): The results of quantum computers promise much higher precision in certain areas than those of classical computers (Havlicek et al. 2018, Sierra-Sosa et al. 2020). As the use of quantitative methods more and more leads to the evaluation of data by means of data analytics, for example techniques from the field of machine learning such as clustering and classification, the quality of the results in terms of precision is becoming increasingly important. Various new approaches, for example for support vector machines (SVM) (Havenstein et al. 2018, Havlicek et al. 2018), or for variational quantum classifier (VQC) (Sierra-Sosa et al. 2020), offer different promising fields of application in digital humanities.

(iv): In addition, the use of quantum computers makes it possible to solve problem classes that were previously considered practically unsolvable (e.g. complexity class BQP (Nielsen and Chuang 2010)). Hereby quantum computers enable approximate solutions for otherwise algorithmically inaccessible problem classes. To what extent and for which fields of applications in the digital humanities this can be beneficially used is still to be explored, but it opens up a whole field of potential questions whose solutions now deem to be possible. A fist promising use case to stress this potential is introduced by (Palmer 2020). Here natural language processing is performed on a quantum computer using the power of quantum states to code complex linguistic structures and novel models of meaning in quantum circuits to understands both, grammatical structure and the meaning of words.

(v): Finally, there are certain types of problems that can only be solved by a quantum computer, i.e. these types of problems can be proven not to be solvable at all by a classical computer (Raz and Tal 2018). Being able to solve new types of problems, the digital humanities have the chance to work on corresponding questions that have not yet been tackled at all, perhaps not even identified or considered. To identify and explore these possible new application areas is an exciting and promising task.

(vi): The use of a quantum computer promises to be significantly cheaper than that of a conventional supercomputer: a price of about 200€ per hour compute time on a quantum computer can be assumed (Dickel 2018). This can be of benefit for the often financially strained humanities, especially for the feasibility of smaller research projects by providing low-cost access to high-performance hardware.

(vii): Furthermore, quantum computers are much more energy efficient than computers with comparable performance (Nielsen and Chuang 2010). Without having a direct influence on digital humanities, the indirect influence should certainly be mentioned. Due to their energy efficiency quantum computers contain a great potential for change, especially in times of climate change.

The above potentials reveal that it is of great importance to develop application knowledge in the domain of quantum computing at an early stage, especially in a field where the mathematical and physical basics required to use this technology cannot be taken for granted.

## 2.3 Current Challenges of Quantum Humanities

The description of the potentials outlined above are from an abstract point of view and must be considered much more comprehensively in its complexity, taking into account various factors from related issues of quantum computing as well as from digital humanities.

From a hardware perspective, the potentials stated above apply to an "ideal" quantum computer, which is not yet available. Today's quantum computers are error-prone and have only limited capabilities (Preskill 2018, Leymann and Barzen 2020a). They provide only a small number of qubits (LaRose 2019) allowing only a limited set of input data to be represented within the quantum computer. Besides this, noise affects calculations on quantum computers (Knill 2007, Preskill 2018): The states of the qubits are only stable for a certain amount of time – an effect referred to as decoherence – due to unintended interactions between the qubits and their environment (Leymann et al. 2020, Nielsen and Chuang 2002). In addition, the operations that manipulate the qubits are also error-prone. Nevertheless, so-called NISQ (Noisy Intermediate-Scale Quantum) machines (Preskill 2018) allow first applications which can be used for prototypical implementations addressing small real-world problems. This can contribute to establish application knowledge – also in the field of quantum humanities. Taking a closer look at the recently published quantum roadmap of IBM (Gambetta 2020) it emphasizes the importance of starting to build application knowledge now: Here IBM outlines its roadmap to 1 million error-corrected qubits within a decade, providing more than 1,000 qubits in 2023, continuously increasing the number of qubits for the coming years.

Next to the issues related to the hardware of quantum computers, from the software perspective too several questions need to be addressed when working with quantum computers. Even though quantum computers are becoming commercially available (e.g. IBM, D-Wave, Rigetti, Honeywell, IONQ), questions regarding usability and accessibility must be considered for each vendor. Also, more concrete questions like the following need to be taken into account: How to encode your data properly based on the processing demands of the chosen quantum algorithm (Weigold et al. 2021)? How to expand oracles that many algorithms contain to keep the influence of such an expansion on the number of qubits and operations required small? How to cope with readout errors? (Leymann and Barzen 2020a).

It is also necessary – even if the quantum computer is superior to classical computers in certain areas as stated above – to address several open questions from the application side. In a first step those existing and new problems from the digital humanities need to be determined that are suitable to be considered being solved by quantum computers. Quantum humanities combines all the different disciplines subsumed under the humanities with disciplines like computer science, mathematics, and physics from the quantum computing site. This makes it an even more heterogenous field than the digital humanities is already, and demands a high degree of translation capabilities. Defining terms that are used differently in the various disciplines to establish a shared language is one of the fundamental tasks for quantum humanities.

Also, when focusing on the more "pragmatical" benefits by using quantum computers as a tool for answering concrete questing from the digital humanities, e.g. how to run parts of a classification algorithm in a machine learning scenario on quantum hardware, lots of questions still remain and need to be answered. For example: How to choose the best algorithm to address the stated problem? How to cope with categorical data while most algorithms require numerical data?

Based on the following use case we want to contribute reusable knowledge for different applications in quantum humanities and introduce a toolchain supporting data analysis by including quantum computers in the process.

## 3 Quantum Humanities Use Case: Project MUSE

In recent years, more and more research projects in the field of digital humanities can be identified that are based on data and data analysis to provide new insights into questions stated in the humanities. For example, when taking a closer look at the books of abstracts of the DHd (Digital Humanites im deutschspachigen Raum (DHd 2021)) conferences – as a well-established conference in the digital humanities – over the last three years (Vogeler 2018, Sahle 2019, Schöch 2020) the term *data* (and its German translation, as well as all compound words containing the term) occurs more than 5,600 times and hint to the significance of data in digital humanities research. Also, in analogy, terms hinting to analyzing this data via techniques from machine learning (e.g. searching for: machine learning, artificial intelligence, un-/supervised learning, clustering and classification) has about 1,000 counts, giving an impression of the importance of this approach to analyze data gained during the last years. Since quantum computers are expected to have substantial superiority especially in the field of machine learning (Schuld and Petruccione 2018), the use case presented in the following is positioned in this area. Here, data from the digital humanities project MUSE is analyzed with the of help machine learning and quantum machine learning techniques.

The project MUSE (MUSE 2021) aims at identifying a pattern language for costumes in films. It contains the method as well as a supporting toolchain to capture all relevant information about costumes, to analyze them and to abstract the significant information into costume patterns. Costume patterns capture those significant elements of a costume (e.g. color, material, way of wearing) that are used to communicate a certain stereotype or a character trait, for example. When speaking of a "nerd" or a "wild west outlaw" most recipients do have a rather clear idea of what these stereotypes should look like to be recognized as "nerd" or "wild west outlaw" based on media socialization. Costume designers compose their characters by choosing each part of a costume, its specifics and attributes with the explicit intent to support the character, the story, and the actor in achieving a specific effect (Schumm et al. 2012, Barzen and Leymann 2014). As there are often similar clothes, colors, or materials used to communicate the same stereotype there are conventions that have been developed and knowledge about these conventions is contained in films.

To extract this knowledge the MUSE method consists of five steps (Barzen et al. 2018, Barzen 2018) which are supported by different tools and are outlined in the following. The method has been proven to be generic by applying it in our parallel project MUSE4Music (Barzen et al. 2016).

### 3.1  MUSE Ontology

Step 1: Define the domain by means of a comprehensive ontology which might be based on several taxonomies (Barzen 2013, Barzen 2018). These taxonomies structure all relevant parameters that have a potential involvement in the effect a costume might have. The MUSE ontology includes taxonomies of base elements (e.g. pants, shirts, jumpers) and primitives (e.g. sleeve types, zipper, collar types), their materials, designs, colors, way of wearing, conditions, and functions – to name just a few – as well as so-called operators (e.g. worn above, wrapped around, attached) turning base elements into a composed outfit. The hierarchical structure of the taxonomies is important when analyzing the data. As various algorithms require numerical data the categorical data of MUSE must be transformed into numerical data based on the structure of the taxonomies, for example (see section 4.2, for more details see (Barzen et al. 2021)). With more than 3150 nodes, a comprehensive ontology to describe costumes in a very detailed and structured manner has been developed.

### 3.2  MUSE Film Corpus

Step 2: Identify – based on well-defined criteria – those films that have a potential "big" impact in terms of costumes on the recipients. Therefore, in a first step genres were determined that promise a reoccurrence of quite similar characters and stereotypes. For MUSE the genres high school comedy, western movies and fairy tales were chosen. In a second step within each of these identified genres the 20 films with the highest box office numbers and scores in rankings were chosen as part of the initial film corpus to be analyzed (for more details see (Barzen 2018)).

### 3.3  MUSE Data Set

Step 3: Capturing detailed information about the costumes of the initial film corpus. To support capturing of all relevant parameters, a repository was designed and implemented: The MUSE-repository (an open-source implementation is available under (MUSE GitHub 2021)) is based on the ontology described in section 3.1 and assists to collect all information about the films of the corpus, the characters and the costumes. Currently (March 2021), the data set contains 5,231 costumes described from 58 films. These costumes contain 28,798 base elements and 63,711 primitives, 159,360 selections of colors and 180,110 selections of materials.

### 3.4 MUSE Data Analysis

Step 4: Analyzing all captured information about costumes to determine those costumes and costume attributes that achieve a similar effect in communicating with the recipient hinting to costume patterns. The analysis consists of two main steps (Falkenthal et al. 2016a): The first step applies data mining techniques, e.g. association rule mining, to determine hypotheses. These hypotheses could, for example, identify those costume elements that are used to communicate a certain character trait like "active person" or "shy person". To refine and verify these hypotheses in the second step, online analytical processing (OLAP) techniques (Falkenthal et al. 2015) are used. As a result, indicators for costume patters are determined.

To improve the process of building hypotheses we are currently extending the analysis of the MUSE data by various techniques from machine learning as well as quantum machine learning (Barzen und Leymann 2020, Barzen et al. 2021). A more detailed discussion on the currently used methods and techniques is given in the following sections 4 to 7.

### 3.5 MUSE Costume Patterns

Step 5: Abstracting the results of the analysis step 4 into costume patterns. Patterns in the tradition of (Alexander et al. 1977) are documents that follow a predefined format to capture knowledge about proven solutions in an abstract way to make this knowledge easily accessible and reusable for other applications. Patterns are related to each other and compose a pattern language based on these relations. As stated above, costume patterns capture the proven solutions about how costume designers address the problem to communicate a certain stereotype like a "wild west sheriff" in terms of all the significant elements of the costume. This contains e.g. base elements, primitives, their relations, colors, ways of wearing, material, if they proved to be significant in the analyze step. As the costume patterns are part of a costume pattern language, they support to solve complex problems by browsing through different solutions related to each other.

To support the accessibility of the costume patterns, tooling support is provided by our generic pattern repository called *Pattern Atlas* ((Leymann and Barzen 2020b), based on our former pattern repository *PatternPedia* (Fehling et al. 2014)).

# 4 Analyzing Data

Analyzing data has two major purposes: discovery and prediction. The collection of techniques focusing primarily on the first purpose is called data mining, the set of techniques focusing primarily on the second purpose is called machine learning. Despite their different primary purposes, data mining and machine learning have a large overlap in terms of the methods they use: Optimization and statistics are core of both approaches, and data mining is even using selective techniques from machine learning. This is why both disciplines are often subsumed by the term data science.

Another difference is how the suitability or appropriateness of a data mining or a machine learning solution is assessed: The appropriateness of a data mining solution is assessed by its capability to discover previously unknown facts, while the appropriateness of a machine learning solution is assessed by correctly reproducing already known knowledge. Once a machine learning solution reproduced known knowledge the solution is considered to correctly deliver previously unknown knowledge.

Despite these differences, the development of a solution based on either of both disciplines follows the same overall procedure which is described in the next section.

## 4.1 Data Analysis Pipeline

In the past, we used data mining technologies to analyze data about costumes in films (see section 3.4., for more details see (Falkenthal et al. 2016a, Barzen 2018)). The general procedure shown in Figure 1 has been used for this analysis but it is also applicable in analyzing data with machine learning techniques (section 5).

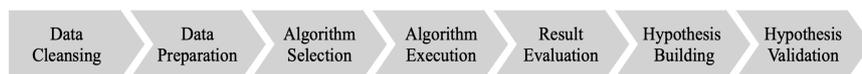

**Fig. 1**. Major steps of the data analysis process chain.

Data cleansing (Skiena 2017) encompasses activities like data acquisition, format conversion and so on. Data preparation (Skiena 2017) deals with a proper treatment and encoding of data, and feature engineering. Algorithm selection (Kubat 2017) determines the family of algorithms (e.g. classification, clustering), selects a proper family member, and choses hyperparameters of the selected algorithm. Result evaluation is often based on visualizing the results (Skiena 2017) and the creation of hypotheses, which are finally validated. Because these last steps are critical for the success of a data analysis project they are depicted as separate steps in Figure 1. If no hypothesis can be built or successfully validated, other algorithms may be tried out to finally succeed with at least one proper hypothesis. Successfully validated hypotheses trigger further processing towards the final goal of the project, like finding patterns (see section 4.3).

Note, that within a data analysis project, most time is typically spend in data cleansing and data preparation (Skiena 2017). Also, several algorithms, even several algorithms of the same family of algorithms, are typically applied within an overall data analysis project. And each algorithm applied involves performing several steps of the process shown in Figure 1. For example, in order to determine patterns of costumes in films in MUSE, the mostly categorial data must be prepared, features must be engineered, and the resulting data must be clustered (see the black shapes in Figure 2). After clusters have been determined, data about newly captured clothes can be classified to identify which costume the clothes represents.

Each of these tasks can be achieved by a variety of algorithms (grey shapes below black shapes). Data preparation, for example, involves one-hot encoding to turn categorial data into binary vectors, or the set of data points representing clothes may be turned into a distance space based on using Wu-Palmer similarity (see section 4.2). Feature engineering may be performed by training an autoencoder (see section 5.5), or multi-dimensional scaling after the transformation of the data into a distance space. Clustering may be based on solving the maximum cut problem (see section 6.2), training a corresponding Boltzmann machine (see section 5.4), or using the k-means algorithm (Arthur and Vassilvitskii 2007). Classification can be done by means of a Boltzmann machine too, or by using a SVM.

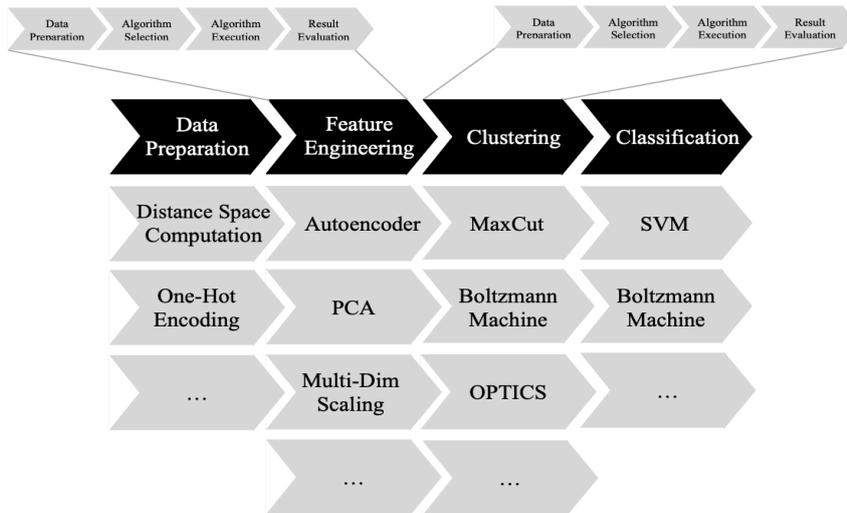

**Fig. 2**. Data analysis pipeline for determining patterns and succeeding classification.

Furthermore, each of the tasks depicted as black shapes in Figure 2 is in fact performed as a process chain (see the small process chains refining two of the black shapes). For example, feature engineering requires to select a corresponding algorithm, e.g. multi-dimensional scaling and setting proper hyperparameters (see section 7.3). The corresponding embedding needs the data in a certain format which has to be prepared. After executing the algorithm, the resulting data representation must be evaluated; if it is not appropriate, the hyperparameters must be adapted, the algorithms must be run again. If the results based on this algorithm is not appropriate at all, another algorithm (e.g. principal component analysis (PCA)) must be tried out. Once the features are properly engineered, clustering is performed, which means that the corresponding process chain is executed. Thus, a *pipeline* results that consists of a sequence of several process chains.

## 4.2 Categorical Data

Many data in the Humanities are categorical data, i.e. data with a finite set of values, often strings, that cannot be used for calculations. For example, computing the maximum of a set of textures or the average of jobs is meaningless. But most machine learning algorithms presume numerical data or even metrical data. Thus, categorical data must be transformed into metrical data to be able to be processed by such algorithms.

(Barzen et al. 2021) discussed a mechanism of how to turn tuples of categorical data the domains of which are tree structured into an approximately equivalent set of vectors in $\mathbb{R}^n$, i.e. into a set of metrical data. Our mechanism is (i) based on the Wu-Palmer similarity (Wu and Palmer 1994), (ii) the fact that similarities can be turned into distance measures, and (iii) that a finite set with a distance measure can be embedded into an appropriate $\mathbb{R}^n$ by means of multidimensional scaling (MDS) (Cox and Cox 2001): see (Barzen et al. 2021) for all the details.

Data elements the values of which are given by means of a taxonomy are such data with a tree structured domain. In our application area of costumes in films (Barzen 2018), most data types have domains defined by taxonomies. Thus, based on our mechanism this data can be embedded into a vector space and, consequently, can be processed by machine learning algorithms.

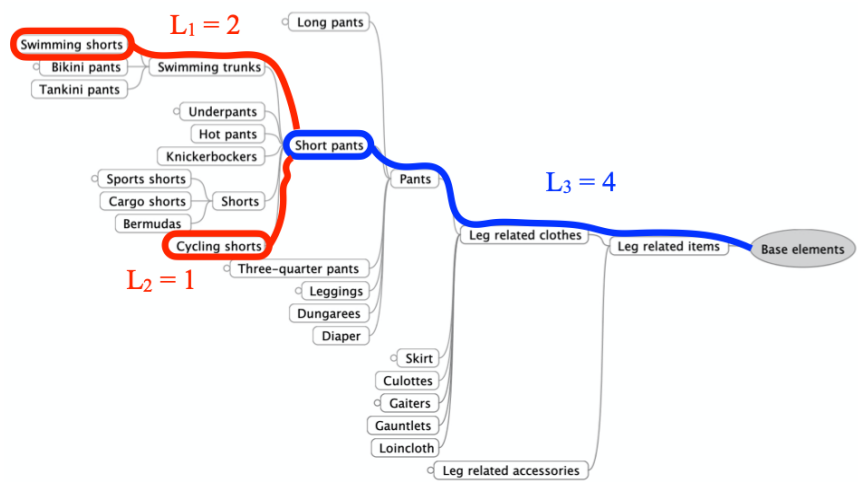

**Fig. 3**. Example for using the Wu-Palmer similarity measure.

Figure 3 shows a small fraction of one of the taxonomies of our costume data set. To compute the similarity of two nodes in a tree (e.g. a "swimming shorts" and a "cycling shorts" in the figure) the lowest common ancestor of these two nodes must be determined: in our example, this is the node "short pants". Next, the length of the paths (in terms of number of edges) from each of these nodes to their lowest common ancestor must be determined: in our example, these lengths are $L_1$=2 and $L_2$=1. Then, the length of the path from their lowest common ancestor to the root of the tree must be determined: in our example, this length is $L_3$=4. Finally, the *Wu-Palmer similarity* is defined as

$$\omega(swimming\ shorts, cycling\ shorts) = \frac{2L_3}{L_1 + L_2 + 2L_3} = \frac{8}{2 + 1 + 8} = \frac{8}{11} \approx 0.72$$

The similarity between two nodes is between 0 and 1, i.e. "swimming shorts" and "cycling shorts" are quite similar. A first quick overview of the similarities between the data elements of a data set can be given by a *similarity matrix* that presents the similarities of pairs of data elements and may render these values by means of a color code (see section 7.1 for an example).

The method to determine the similarity ω between two values of a tree structured domain (part 1 in Figure 4) can be extended to determine the similarity of two sets of values of the same tree structured domain: in a nutshell, the similarities ω of all pairs is determined whose first component is from the first set and whose second component is from the second set; a certain mean value of these similarities is computed, resulting in a similarity measure σ between two sets (part 2 of Figure 4). Based on the similarity measure σ the similarity of tuples of the same dimension can be determined whose components are sets of values of the same tree structured domain (part 3 in Figure 4): the similarities between each components of the tuples is determined (e.g. the similarity of the colors of two tuples representing costume A and costume B, the similarity of their materials and so on); the mean value of these similarities is computed which results in a similarity measure μ between two tuples. See (Barzen et al. 2021) for the details.

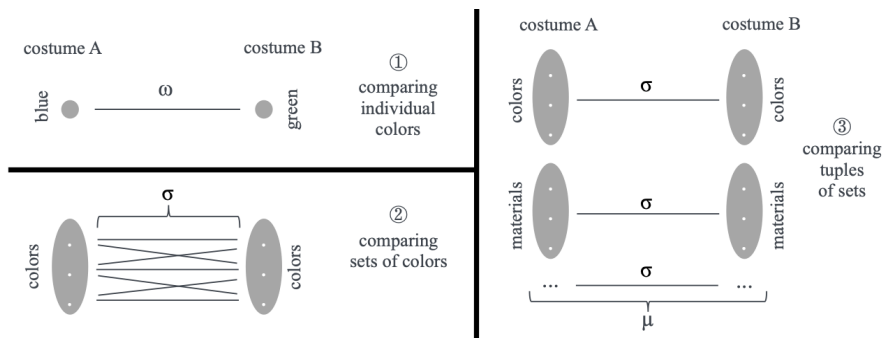

**Fig. 4.** Computing similarities of sets and tuples.

The similarity ω(i,j) between two nodes i and j (and the similarity σ(i,j) of two sets or the similarity μ(i,j) between two tuples i and j, respectively) can be transformed into the dissimilarity or *distance* δ(i,j) between i and j, respectively, by means of

$$\delta(i,j) = \sqrt{\omega(i,i) + \omega(j,j) - 2\omega(i,j)},$$

with ω(i,i) = ω(j,j) = 1 (substituting ω by σ or μ in the formula accordingly). This way, a set of data points the tuples of which have components whose domains define by taxonomies can be transformed into a set with a distance measure. In analogy to a similarity matrix the distances between the data elements are used to build a *distance matrix*.

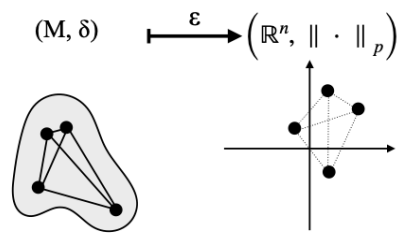

**Fig. 5.** An embedding maps a set with a distance measure into a Euclidian vector space.

Figure 5 depicts this situation: A set M with a distance measure δ is given, and the distances of the elements of M are indicated by the length of the lines between the points. An *embedding* ε maps the set M into $\mathbb{R}^n$, i.e. the elements of M become vectors in $\mathbb{R}^n$ (often referred to as *feature space*). But the vectors are not chosen arbitrarily but the mapping ε tries to keep the distances in M approximately the same than the distances in $\mathbb{R}^n$, i.e. the distances of the corresponding vectors (indicated by the length of the dashed lines between the vectors in $\mathbb{R}^n$) are approximately the same as the distances in M; the distances in $\mathbb{R}^n$ are measured by some norm $\|\cdot\|_p$. Here, "n" and "p" are hyperparameters that must be chosen based on former experience. Embeddings can be computed by multidimensional scaling (MDS) (Cox and Cox 2001).

This way, a set of data points M with components whose domains are defined by taxonomies can be transformed into a set of data points in $\mathbb{R}^n$ such that the distances in M are approximately the distances in $\mathbb{R}^n$. After this transformation, the data points can be further processed by most machine learning algorithms. Our costume data set is transformed in this manner (see section 7.1) and is, thus, available to be analyzed by several machine learning algorithms.

### 4.3 Creating Pattern Languages Based on Data Analysis

The main purpose of our project MUSE is the creation of pattern languages of costumes in films. Pattern languages can be built in several ways. Often, they are built from experience: practitioners recognize that a particular problem occurred repeatedly in the past perhaps in various contexts, and that the solutions applied to solve the problem are very similar. Thus, these solutions can be abstracted to identify their underlying principles such that these principles can be applied to solve that particular problem even in new contexts. This way, a new pattern has been identified. Similarly, relations between patterns are established based on experience. Those relations between patterns become links with an associated semantics.

What results is a pattern language: A *pattern language* is a collection of patterns within a particular domain and their relations. From an abstract point of view, a pattern language is a weighted, directed graph. The nodes of this graph are the patterns, the edges are the links between the patterns, and the weights of the edges are the semantics of the links.

In contrast to derive patterns from experience, our project MUSE strives towards deriving patterns by means of analyzing data. The method we developed in MUSE is generally applicable and its use has already been initially verified also in the domain of music (Barzen et al. 2016) and is envisioned by other aspects of films (see Falkenthal et al. 2016b). Applying the MUSE method in other areas of the humanities successfully seems to be possible. After having used data mining techniques in our method first, we are now exploiting machine learning techniques and the initial results we get seem to be more promising than using mining techniques.

To identify a pattern language based on data analysis the process sketched in Figure 1 in section 4.1 must be extended (see Figure 6). If the hypothesis validation step resulted in a verified hypothesis, this hypothesis must be turned into a pattern document. I.e. the various sections of the pattern document corresponding to the hypothesis must be authored. Thus, a pattern candidate is created: the resulting document is "only" a candidate but not yet a pattern because its general applicability still needs to be verified. Such a verification is performed in the succeeding pattern verification step. (Reiners 2014) proposes a system based on which a community e.g. can discuss the candidate, exchange experiences with it, and can jointly decide to promote the candidate to a pattern. The new pattern must be related to the other patterns of the domain in the pattern relations step. This way a pattern language results. Note, that a pattern language is a "living" entity, i.e. over the time patterns will be added to the pattern language. Typically, a first version of a pattern language will be made available or published when "major" problems of the underlying domain can be solved with it.

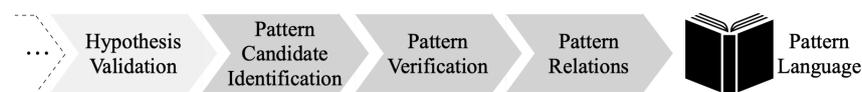

**Fig. 6**. Creating a pattern language.

# 5 Artificial Neural Networks

An artificial neural network (or neural network for short) is a mathematical model that mimics the biological neural networks of brains of animals. Such a neural network is represented as a directed graph the nodes of which abstract neurons and the edges abstract synapses of a brain. In this section, neurons and neural networks are defined, perceptrons and their use for classification are sketched, and restricted Boltzmann machines, as well as autoencoders are described.

## 5.1 Neurons

The first model of an artificial neuron roots back to (McCulloch and Pitts 1943), nowadays called a McCulloch-Pitts neuron. Such a neuron accepts binary data as input and produces a binary output if a predefined threshold is exceeded. This model was extended and called perceptron (see section 5.3) in (Rosenblatt 1958). Today a neuron is abstracted as a mathematical function (Zurada 1992): it receives multiple values as input and produces a single value as output. What characterizes functions that represent neurons is the way in which the output of a neuron is computed, i.e. the ingredients of such functions and how they interact.

**<u>Definition</u>**: Let $w_1, \dots, w_n \in \mathbb{R}$ be real numbers called *weights*; the function

$$\Pi \colon \mathbb{R}^n \to \mathbb{R}, \ x \mapsto \sum_{i=1}^{n} w_i x_i$$

is called *propagation function*. Furthermore, a function $\alpha \colon \mathbb{R} \to \mathbb{R}$ is called *activation function*. Then, a *neuron* is a map

$$\nu \colon \mathbb{R}^n \to \mathbb{R}, \ x \mapsto \alpha\left(\sum_{i=1}^{n} w_i x_i\right)$$

∎

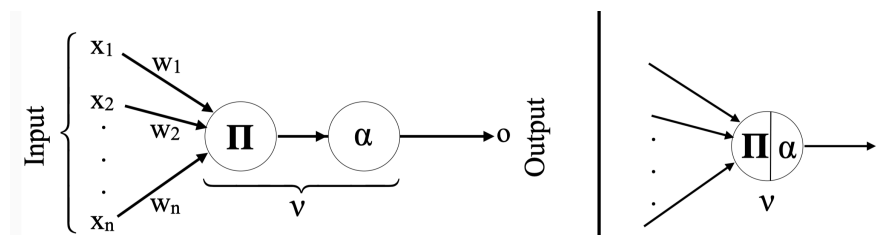

**Fig. 7**. Ingredients of an artificial neuron.

Figure 7 shows a graphical representation of an artificial neuron: the left side of the figure represents the details of the definition before, the right side of the figure depicts a more condensed version of this graphical representation in which the two nodes representing $\Pi$ and $\alpha$ are combined into a single node denoted with $\Pi|\alpha$. The input values $x_1, \dots, x_n$ of the neuron $\nu$ are connected via directed edges with the neuron's propagation function $\Pi$; the weight $w_i$ of the input $x_i$ is shown as a number associated with the corresponding edge. The purpose of the propagation function $\Pi$ is to aggregate the input of the neuron considering the corresponding weights of the input values. The purpose of the activation function $\alpha$ is (i) to decide whether its input suffice to pass on a non-zero output and (ii) what the output value will be.

An important variant defines a neuron as a map $\nu: \{1\} \times \mathbb{R}^n \subseteq \mathbb{R}^{n+1} \to \mathbb{R}$, with

$$\nu(1, x_1, \ldots, x_n) = \alpha \left( w_0 + \sum_{i=1}^{n} w_i x_i \right).$$

Thus, a neuron has one input $x_0$ of constant value "1" (see Figure 8). The corresponding weight $w_0$ is called *bias*, denoted as "b". The bias influences the firing of the neuron: a negative bias results in the neuron firing less frequently, a positive bias results in the neuron firing more frequently.

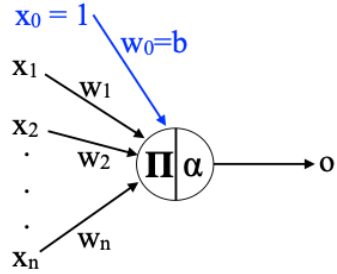

**Fig. 8**. A neuron with a bias.

One problem with realizing neurons on a quantum computer is that quantum algorithms are unitary transformation, thus, linear, but activation functions of neurons are typically non-linear. (Cao et al. 2017) and (Wan et al. 2017) present quantum neurons with different activation functions. A special neuron (a so-called perceptron - see section 5.3) on a quantum computer of the IBM Quantum Experience family has been described in (Tacchino et al. 2018). This implementation proves the exponential advantage of using quantum computers in terms of requiring only N qubits in order to represent input bit-strings of $2^N$ bits. But in order to prepare this input an exponential number of 1-qubit gates and CNOTs are needed, in general (Leymann and Barzen 2020a).

## 5.2 Neural Networks

Neural networks are directed graphs the nodes of which are neurons (Kubat 2017, Skiena 2017). The edges of the graph specify which neuron passes its output as input to which other neuron. The neurons are partitioned into disjoint sets $L_1, \ldots, L_n$ (so-called layers) such that neurons of layer $L_i$ are connected with neurons of layer $L_{i+1}$. The neurons of layer $L_1$ get their input from the outside, i.e. not from other neurons; this layer is called the input layer of the neural net. Neurons of layer $L_n$ pass their output to the outside, i.e. not to other neurons; this layer is called the output layer of the neural network. All other layers are referred to as hidden layers. A neural network with at least one hidden layer is sometimes called a deep neural network.

__Definition__: A *neural network* is a tuple (N, E, $\mathscr{L}$, X, Y) with the following properties:

- G = (N, E) is a directed graph,
- $\mathscr{L}$ = {$L_1$,…,$L_n$} ⊆ $\mathcal{P}$(N) is a partition of the set of nodes, i.e. ∀i: $L_i \neq \varnothing$, $L_i \cap L_j = \varnothing$ for i≠j, and $\cup L_i$=N; $L_i$ is called *layer*, $L_1$ *input layer*, $L_n$ *output layer*, and $L_i$ (1<i<n) *hidden layer*,
- each node $v_i = (\{w_{ji}\}, \Pi_i, \alpha_i)$ ∈N is a neuron with a set of weights {$w_{ji}$}, a propagation function $\mathbf{\Pi}_i$, and an activation function $\alpha_i$,
- X = {$x_1$,…,$x_m$} is the set of *input values*,
- Y = {$y_1$,…,$y_k$} is the set of *output values*,
- the set of edges E ⊆ (N×N) ∪ (X×N) ∪ (N×Y) connects two neurons, or an input value and a neuron, or a neuron and an output value,
- for $v \in L_i$ and (v, v') ∈E∩(N×N) it is v' ∈ $L_{i+1}$,
- for $v \in L_i$ , x ∈ X, and (x, v) ∈ E it is i=1,
- for $v \in L_i$ , y ∈ Y, and (v, y) ∈ E it is i=n, and y is the output of v,
- for ($v_j$, $v_i$) ∈ E∩(N×N) the output of $v_j$ is passed to $v_i$ where it is processed by $v_i$'s propagation function $\mathbf{\Pi}_i$ weighted by $w_{ji}$,
- for $v \in L_1$ and (x, v) ∈ E, v processes x by its propagation function $\mathbf{\Pi}$ weighted by $w_{ji}$.

∎

Effectively, a neural network that consumes m input values and that produces k output values is a function $\mathbb{R}^m \to \mathbb{R}^k$. The computational power of neural networks stems from the fact, that any continuous function on a compact set $K \supseteq \mathbb{R}^m \to \mathbb{R}^k$ can be approximated with arbitrary precision by a neural network (see (Kidger and Lyons 2020) for details).

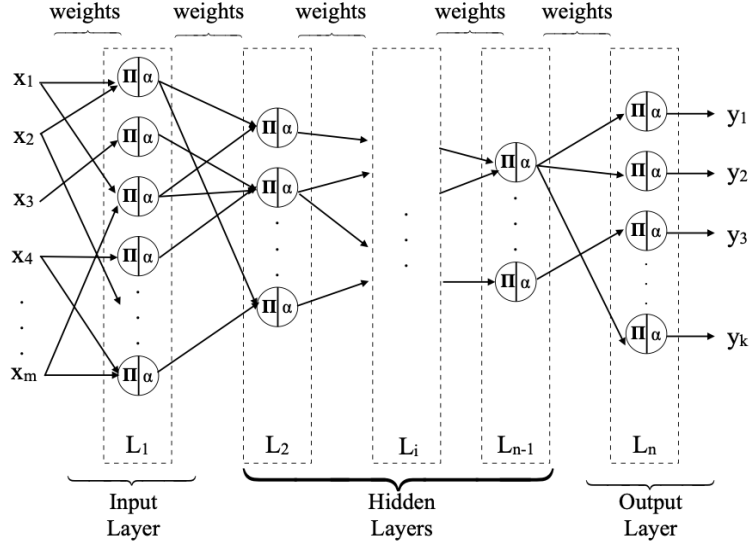

**Fig. 9**. Principle structure of a neural network.

Figure 9 shows the principle structure of a neural network according to our definition before. Note, that there are some variants of our definition: sometimes, edges between neurons with the same layer are supported, sometimes edges are allowed between neurons of any layer, sometimes a neuron may be connected to itself, and so an. However, such variations are for use in special cases.

The structure of a neural network, i.e. its set of layers and its edges, is referred to as its *topology*. Defining the topology of a neural network, specifying the activation functions and propagation functions of its neurons is called *modeling* a neural network. Note, that most often the propagation function from our definition in section 5.1 is used, but activation functions change across problem domains; even different activation functions for different neurons of a given neural network may be defined.

Training a neural network consists of choosing appropriate weights and biases. For this purpose, the neural network is considered as a function $F: \mathbb{R}^m \to \mathbb{R}^k$. Furthermore, a set of input vectors $\{x_j\} \in \mathbb{R}^m$ is given together with the corresponding known correct output vectors $\{y_j\} \in \mathbb{R}^k$; note, that the set of pairs $\{(x_j, y_j)\}$ is called *training data*. Ideally, the neural network will output $y_j$ for each input $x_j$; but the neural network is not (yet) realizing the ideal function, i.e. it is just approximating it: it will output $Y_j = F(x_j)$. Thus, the neural network will produce an error $|y_j - Y_j|$ in computing $y_j$. The goal of training the neural network is to minimize these errors considering all training data by adapting the weights and biases.

For this purpose, the so-called *loss function*

$$L(w, b) = \sum_j \left(y_j - Y_j\right)^2 = \sum_j \left(y_j - F(x_j)\right)^2$$

is minimized (as usual, the square of errors is used instead of their modulus). Here, "w" denotes all weights of the neural network and "b" all its biases. *Training* now means to choose w and b such that L(w,b) is minimized. Lots of mathematical procedures are known to minimize L, e.g. gradient-based methods (e.g. stochastic gradient descent) or derivative-free methods (e.g. Nelder-Mead) can be chosen (Nocedal and Wright 2006). Note, that gradient-based methods require differentiability of the loss function which in turn requires differentiability of F which in turn requires differentiability of the activation functions of the neurons composing the neural network implementing F (see section 5.1). After this training phase, the neural network is ready for use. It is said that the neural network has *learned*. The use of training data, i.e. input values with corresponding known output values, is referred to as *supervised learning*: the processing of the input data by the neural network is supervised by comparing its output with the known given results associated with the input. Learning without supervision is referred to as *unsupervised learning*: this kind of learning is discussed in section 5.5 in the context of autoencoders.

An implementation of quantum neural network that requires a single qubit for each neuron (plus additional ancillae) has been proposed by (Cao et al. 2017). The training of this quantum neural network is performed in a hybrid quantum-classical manner, i.e. the optimization is executed by classical software using the Nelder-Mead algorithm. In contrast to training classical neural networks where the individual training data is processed sequentially, the training of this quantum neural network can be done based on a superposition of the input/output pairs of the training data - something impossible for classical neural networks. A quantum neural network has been realized on a near-term quantum processor as described by (Farhi and Neven 2018); their neural network has been successfully used as a binary classifier on the MNIST data set to distinguish two different handwritten digits. A set of requirements on implementations of quantum algorithms that represent "meaningful" quantum neural networks has been posed in (Schuld et al. 2014).

## 5.3 Perceptrons

A perceptron can decide whether a given data point represented as input is left or right of a given straight line in the plane (or in higher dimensions, right or left of a hyperplane). Thus, a perceptron can be used as a binary classifier of linear separable data sets: the left side of Figure 10 shows two kinds of data points (grey points and black points) in the plane. They are linear separable because a straight line H can be found that splits these two data sets such that on one side of H only the black data points are located, on the other side only gray data points reside.

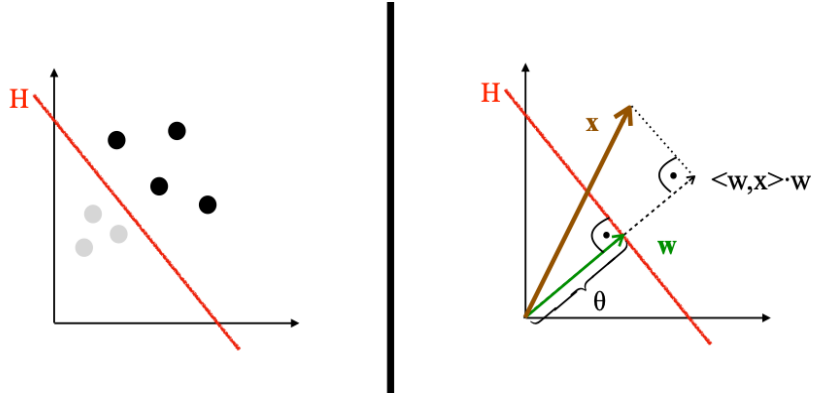

**Fig. 10**. Binary classifier for linear separable sets.

Any straight line H (or hyperplane in higher dimensions) can be defined by means of a vector w that is orthogonal to H, and whose length $\parallel w \parallel = \theta$ is the distance of H from the origin (so-called Hesse normal form - see the right side of the figure). For any point x, the scalar product $\langle w, x \rangle$ is the length of the projection of x onto w, and $\langle w, x \rangle \cdot x$ itself is the projection of x onto the straight line defined by w. Thus, a point x is on H if and only if $\langle w, x \rangle - \theta = 0$. Furthermore, x is "on the right side" of H if the length $\langle w, x \rangle$ of the projection of x onto w is greater than $\theta$ (i.e. $\langle w, x \rangle - \theta > 0$), and x is "on the left side" of H if the length $\langle w, x \rangle$ of the projection of x onto w is less than $\theta$ (i.e. $\langle w, x \rangle - \theta < 0$). Consequently, the value of $\langle w, x \rangle - \theta$ determines whether a point x is right or left of the hyperplane H defined by w, i.e. it serves as is a binary classifier: a positive value classifies x as a black point, a negative value classifies x as a gray point.

**Definition**: A *perceptron* is a neuron ν with Heaviside-function $\Phi_\theta$ as activation function, where $\theta := \parallel (w_1, \dots, w_n) \parallel$ for the weights $w_1, \dots, w_n$ of the neuron.

∎

The vector $w = (w_1, \dots, w_n)$ of the weights of the perceptron ν determines a hyperplane H (with $\parallel w \parallel = \theta$). When a point $x = (x_1, \dots, x_n)$ is passed as input to the propagation function **Π** of ν, the scalar product of w and x results:

$$\Pi(x) = \sum_{i=1}^{n} w_i x_i = \langle w, x \rangle$$

With

$$\nu(x) = \Phi_\theta \left( \sum_{i=1}^{n} w_i x_i \right) \in \{0,1\}$$

it is $\nu(x) = 1 \Leftrightarrow \Phi_\theta(\langle w, x \rangle) = 1 \Leftrightarrow \langle w, x \rangle \geq \theta \Leftrightarrow \langle w, x \rangle - \theta \geq 0$. Thus, $\nu(x) = 1$ classifies x as "right of H" (considering x on H, i.e. $\langle w, x \rangle - \theta = 0$, as "right of H"), and $\nu(x) = 0$ classifies x as "left of H". This reveals that a perceptron is a binary classifier.

Given a linear separable set of data, how are the weights $w = (w_1, \dots, w_n)$ determined the corresponding hyperplane of which splits the data set into two disjoint parts? In other words: how does a perceptron learn to split the data set correctly? For this purpose, the following procedure is used (e.g. (da Silva et al. 2017)):

- Assume a set $T = \{(t_1, c_1), \cdots, (t_m, c_m)\}$ of training data is given, i.e. $t_j = (1, x_{1j}, \cdots, x_{nj})$ is a data point (with its first component set to "1" to introduce its weight as a bias) and $c_j \in \{0,1\}$ classifies the data point as member of the class "0" or member of the class "1". E.g. the class "0" may indicate a grey point and the class "1" a black point in Figure 10. Since the training data specifies the correct result of the classification, the algorithm sketched is a supervised learning algorithm.

- The algorithm is performed in several steps $\tau$; the tuple $w(\tau) = (w_0(\tau), w_1(\tau), \cdots, w_n(\tau))$ denotes the bias $b(\tau) = w_0(\tau)$ and the weights of the perceptron at step $\tau$.

- At step $\tau = 0$, w(0) is chosen randomly ("small" values have been shown to be a good first choice).

- At any step $\tau > 0$ the following is computed for $1 \leq j \leq m$, i.e. for the whole set of training data:

$$y_j(\tau) = \Phi_{b(\tau)}\left(\sum_{i=1}^{n} w_i(\tau) x_{ij} + b(\tau)\right) = \Phi_{b(\tau)}\big(\langle w(\tau), t_j \rangle\big) \in \{0,1\},$$

  i.e. $y_j(\tau)$ is the classification of the data point $t_j$ at step $\tau$.

- The value $\left| c_j - y_j(\tau) \right|$ measures the error of classifying the data point $t_j$ at step $\tau$ as $y_j(\tau)$ while $c_j$ is the correct known classification. The goal is to minimize the overall classification error

$$e(\tau) = \frac{1}{m} \sum_{k=1}^{m} |d_k - y_k(\tau)|$$

  i.e. e($\tau$) < $\gamma$, where $\gamma$ is a predetermined error threshold. If the error is below this threshold the algorithms stops.

- Otherwise, a new bias and new weights are determined: chose an arbitrary data point $t_j$ (i.e. $1 \leq j \leq$ m) and compute

$$w_i(\tau + 1) = w_i(\tau) + r \cdot \left( c_j - y_j(\tau) \right) \cdot x_{ij}$$

  for $0 \leq i \leq$ n. Thus, if the computed classification $y_j(\tau)$ for data point $t_j$ is correct, i.e. it is the known classification $c_j$, $w(\tau)$ is not changed "in direction of $t_j$", i.e. $w(\tau + 1) = w(\tau)$. Otherwise, $w(\tau)$ is modified "in direction of $t_j$", i.e. $w(\tau + 1) = w(\tau) \pm r \cdot t_j$. Here, r is a predetermined constant, the so-called "learning rate".

Assuming linear separable training data, the above algorithm converges and separates the training data correctly (Novikoff 1962). Typically, the algorithm converges fast, i.e. the condition $e(\tau) < \gamma$ is met fast. But there are situations in which the convergence is slow and the algorithm should be stopped even if the result is not quite precise. This is covered by the additional condition "$\tau = N$" limiting the number of iterations even if $e(\tau) \geq \gamma$, i.e. the result of the learning process is above the error threshold. Note, that the variables $\gamma$, r, and N are so-called *hyperparameters* and must be properly set prior to the learning process, e.g. based on former experience.

The separating hyperplane learned by the perceptron depends on the initial values of the bias and the weights $w(0)$ chosen during the initial step $\tau=0$. This is depicted on the left side of Figure 11: different initial values $w(0)$ result in different hyperplanes $H_1$, $H_2$, $H_3$. These hyperplanes are differently suited in correctly classifying new data points: two new data points are shown as a triangle and a square. The triangle is classified as a black point by $H_1$ and $H_3$ (being "right" of the hyperplanes), and as a grey point by $H_2$ (being left of it). The square is classified by $H_3$ as a black point, but as a grey point by $H_1$ and $H_2$. This non-determinism of a perceptron learning a separating hyperplane is in contrast to another machine learning technique (which is not based on neural networks at all): a support vector machine (SVM) (Burges 1998). A support vector machine determines a unique separating hyperplane based on linear separable training data: this is achieved by computing the hyperplane with the maximum margin that does not contain any of the test data (see the right side of Figure 11 - the grey shaded rectangle represents the margin of the hyperplane). A support vector machine would classify the square as a grey point, and the triangle as a black point. Furthermore, support vector machines can even be used to classify non-linear separable data sets (Bennett and Campbell 2000) by embedding the data into a high-dimensional vector space (which already hints that quantum implementations of support vector machines are advantageous because the state spaces of quantum computers are extremely huge).

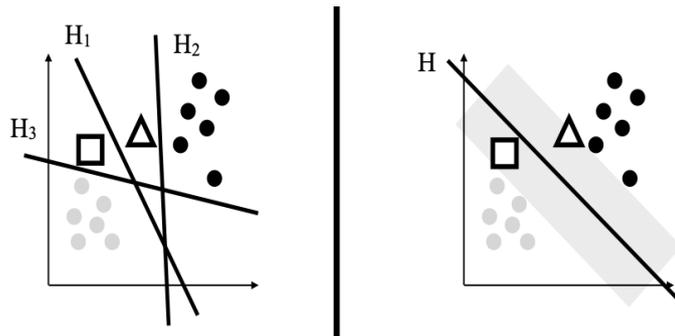

**Fig. 11.** Computing separating hyperplanes via perceptrons and support vector machines, and using a computed hyperplane as classifier for new data.

(Wiebe et al. 2016) proposes two quantum perceptrons: the first one can be trained with a quadratic speedup in size of the training data, the second one quadratically reduces the training time required to achieve a given precision when compared with the algorithm presented in this section. A quantum perceptron has been suggested in (Tacchino et al. 2018) that requires exponentially less storage resources for training; it has been implemented and validated on a near-term quantum computer of the IBM Q family of systems.

### 5.4 Restricted Boltzmann Machines

Restricted Boltzmann machines (RBMs) are attributed to (Smolensky 1986). While for general Boltzmann machines learning is not feasible under practical conditions, restricted Boltzmann machines can learn efficiently (see (Hinton 2012) for an overview). A restricted Boltzmann machine is a neural network with two layers: an input layer and a hidden layer (see Figure 12). Each neuron of the input layer has a single input value, and the weight associated with such an input is "1". Input values as well as the output of each neuron are Boolean values; especially the output of a restricted Boltzmann machine, i.e. the output of the neurons of the hidden layer, are Boolean values. Each neuron has a bias (not shown in the Figure 12).

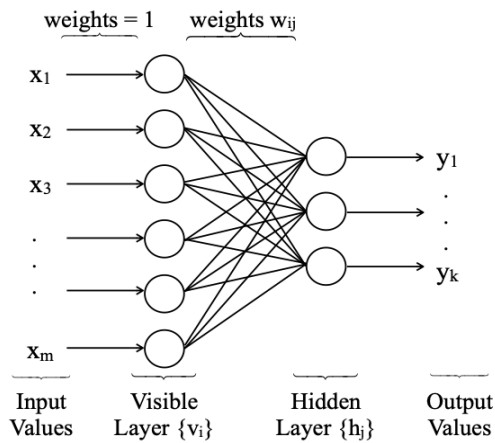

**Fig. 12**. Structure of a restricted Boltzmann machine; biases are not shown.

<u>**Definition**</u>: A *restricted Boltzmann machine* is a two layer neural network (N, E, $\mathscr{L}$, X, Y) with following properties:

- the output of each neuron is a Boolean value (especially, $y \in \{0, 1\}$ for $y \in Y$),

- the input values are Boolean values, i.e. $x \in \{0, 1\}$ for $x \in X$,

- each neuron $v_i \in N$ has a bias; for $v_i \in L_1$ the bias is denoted by $a_i$, for $v_i \in L_2$ the bias is denoted by $b_i$,

- for $v_i \in L_1$ there exists exactly one $x_i \in X$ with $(x_i, v_i) \in E$ (ignoring the input "1" for the bias),
- for $v_i \in L_1$ the weight $w_{ii}$ of $(x_i, v_i) \in E$ is "1",
- $L_1 \times L_2 \subseteq E$.

$L_1$ is called *visible layer*, $L_2$ is called *hidden layer,* i.e. the last property enforces that each neuron of the visible layer is connected to each neuron of the hidden layer.

∎

Training of a restricted Boltzmann machine ((Hinton 2012) for all details) is different from the general mechanism for training a neural network (see section 5.2). The input values of a restricted Boltzmann machine are Boolean values. But data to be processed is typically not Boolean. For this purpose, each neuron $v$ of the input layer represents an observed value $o$: the value $o$ is observed if and only if the input of $v$ is "1". Especially, $o$ maybe of any type of data, i.e. not only numerical data but also categorical data. This way, restricted Boltzmann machines can be used with any type of data as input; and similar for output data.

The application areas of restricted Boltzmann machines include, for example, classification (e.g. (Chen and Srihari 2015, Larochelle et al. 2012)) and feature learning (e.g. (Zheng et al. 2013, Tomczak 2016)). For classification purposes, the training data $T = \{(t, c)\}$ are one-hot encoded, i.e. the class indicator $c$ becomes a vector $c = (k_1, \ldots, k_r) \in \{0,1\}^r$ (where r is the number of classes) with $k_j = 1$ if t is of class j, and $k_i = 0$ otherwise (see the left side of Figure 13). Note, that this is multi-class classification, i.e. data can be associated with more than two classes. For feature learning (right side of Figure 13), the properties of the data correspond to the input and the output are the features learned. Typically, these features are used by another algorithm for further processing, e.g. by another restricted Boltzmann machine (so-called "stacked" restricted Boltzmann machines).

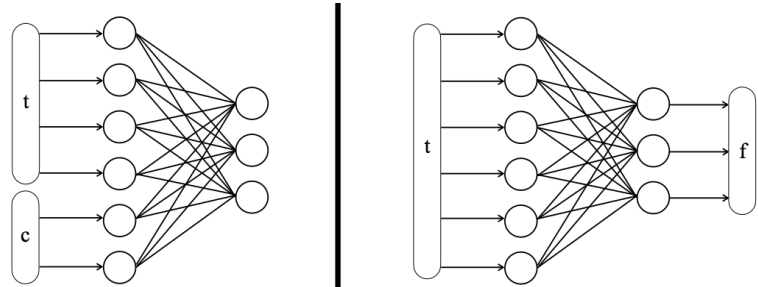

**Fig. 13**. Using restricted Boltzmann machines for classification and feature learning.

A special kind of quantum restricted Boltzmann machines that can approximate any quantum circuit has been specified in (Wu et al. 2020). Also, an implementation of such kind of a quantum restricted Boltzmann machine on a NISQ device of the IBM Q family has been provided. Quantum Boltzmann machines (i.e. a general

Boltzmann machine which supports also connections between neurons of the same layer) as well as quantum restricted Boltzmann machines have been investigated by (Amin et al. 2018) with the result, that their quantum Boltzmann machine outperforms a classical Boltzmann machine on a specific learning task. A circuit realizing a quantum restricted Boltzmann machine is given in (Zhang and Zhou 2015). No implementation on a real quantum computer has been evaluated but a simulation: it turned out that the quantum restricted Boltzmann machine performed a classification task faster and with higher precision than a classical machine. An implementation of a restricted Boltzmann machine to be realized on a D-Wave quantum annealer is described in (Denil and Freitas 2011); obstructions for a real implementation are discussed.

### 5.5 Autoencoders

Today's concept of an autoencoder goes back to (Hinton and Zemel 1994). At a first glance, it can be considered as two restricted Boltzmann machines merged together the purpose of which is to reconstruct their input as output. After training, the autoencoder may be split again into two restricted Boltzmann machines, each of which may be used in combination with further machine learning algorithms.

**Definition**: An *autoencoder* is a three layer neural network (N, E, $\mathscr{S}$, X, Y) with following properties:
- let F be the function represented by the autoencoder, then: $F \approx id$
  (i.e. the autoencoder's output approximates its input),
- card $L_1$ = card $L_3$
  (i.e. input layer and output layer contain the same number of neurons),

The hidden layer $L_2$ is called *code (layer)*. Input layer and code layer are collectively referred to as *encoder*, code layer and output layer are collectively referred to as *decoder*.

∎

Note, that there are variants of autoencoders that support more than one hidden layer, which is not discussed here. Figure 14 depicts on its left side the principle structure of an autoencoder consisting of an encoder and a decoder. The right side of the figure shows its structure as a neural network with three layers, the hidden layer named "code". Roughly, it can be considered as two restricted Boltzmann machines with the code layer being the hidden layer of the encoder and the input layer of the decoder ("stacked machines").

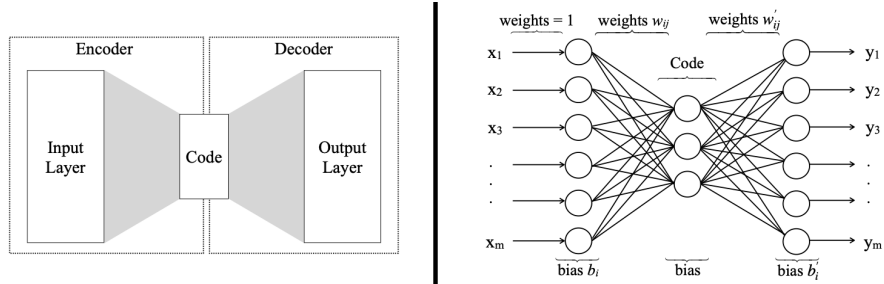

**Fig. 14**. Principle structure of an autoencoder and its structure as a neural network.

The encoder transforms its input into a code (also called a *representation* of the input). If the input is a data point in $\mathbb{R}^m$ and the code is a data point in $\mathbb{R}^k$, the encoder is a map $\varphi \colon \mathbb{R}^m \to \mathbb{R}^k$. Because the output layer must have the same number of neurons as the input layer, the output is again a data point in $\mathbb{R}^m$, i.e. the decoder is a map $\psi \colon \mathbb{R}^k \to \mathbb{R}^m$. By definition, it is $\psi \circ \varphi \approx id$. I.e., for any input x, the output is $\psi \circ \varphi(x) \approx id(x) = x$; thus, the difference (or error) between x and $\psi \circ \varphi(x)$ should be as small as possible (as usual the square of the error $\left(x - \psi \circ \varphi(x)\right)^2$ is taken). For a set of training data $T = \{x^k\}$, the sum of these squared errors should be minimal. Now, $\varphi$ and $\psi$ perform their mapping based on their weights $w_{ij}$ and $w_{ij}'$ as well as their biases $b_i$ and $b_i'$, respectively. Consequently, the sum of the squared errors is a function L ("loss function") of these weights and biases:

$$L(w, b) = \sum_k \left(x^k - \psi \circ \varphi(x^k)\right)^2$$

Training an autoencoder means minimizing this loss function based on some optimization algorithm, i.e. to determine $\left\{w_{ij}, b_i, w_{ij}', b_i'\right\}$ such that $L\left(\left\{w_{ij}, b_i, w_{ij}', b_i'\right\}\right) \approx 0$. Note, that the training of an autoencoder is unsupervised because no set of training data is required that is labeled to indicate the output for a given input.

After training, the encoder may be split from the decoder. The encoder will then transform a data point $x \in \mathbb{R}^m$ into a data point $\varphi(x) \in \mathbb{R}^k$, and it is known that this $\varphi(\mathrm{x})$ represents x faithful, i.e. it contains "the essence of x" because it includes all information to reconstruct x (by means of $\psi$): $\varphi(\mathrm{x})$ represents the features of x. This is the reason for using the term "feature learning" for training an encoder. If k<m, i.e. if the code has less components than the input, a dimension reduction is achieved. Reducing the dimension of data is key because many algorithms are very sensitive to it, i.e. they are much more efficient if a data point to be processed has less components.

Note the similarity to embeddings as discussed in section 4.2 which also reduce dimensions of data. In contrast to encoders, embeddings try to keep distances between data points intact. While some other algorithms that might be used to process data after dimension reduction require similar distances between the high-dimensional data points and the low-dimensional data points, some other do not consider these distances. Thus, the next algorithm used to process the data influences the decision whether an embedding has to be used or whether an encoder suffice.

Based on the work of (Lamata et al. 2019), a quantum autoencoder requiring for each neuron a separate qubit has been worked out in (Bondarenko and Feldmann 2020). Each neuron has a unitary operation associated with it that operates on the neuron and all neurons of the preceding layer. The number of such operations needed to apply the autoencoder grows exponentially with the maximum number of neurons per layer. Training of their autoencoder has been simulated in MATLAB. An implementation of a (small) autoencoder on a Rigetti Quantum Computer has been described in (Ding et al. 2019); the corresponding circuits are provided. This autoencoder was successfully used for dimension reduction. Similar has been achieved in (Pepper et al. 2019) based on a photonic quantum computer. (Romero et al. 2017) describes a variational quantum-classical way to implement a quantum autoencoder. Figure 15 depicts (for illustration only, not for reading) our implementation of the quantum part of the algorithm based on the simulator of IBM Q und Qiskit (Qiskit 2021). The input data is prepared in four qubits that are transformed by the encoder. Once the encoder finished its processing, the first two qubits are set to $|0\rangle$ while the last two qubits are kept in the state the encoder produced. Effectively, the code of the autoencoder consists of these two last qubits, i.e. a dimension reduction by 50% is achieved. Next, the decoder tries to reconstruct the original input from the code. The decoding result is measured and analyzed by a classical part (not shown in the figure); based on this analysis another iteration may be required. Once the autoencoder is successfully trained, the encoder can be used for dimension reduction.

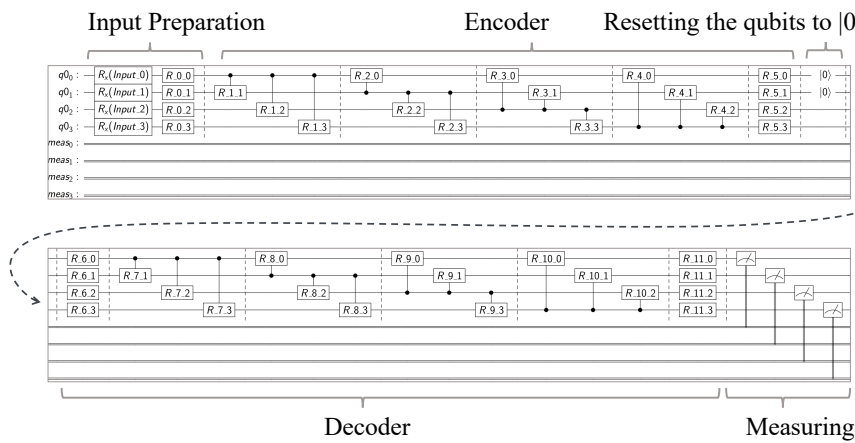

**Fig. 15**. Principle structure of an autoencoder and its structure as a neural network.

# 6 Variational Hybrid Quantum-Classical Algorithms

As described in section 2.3, near-term quantum computers support "short" computations only due to decoherence of their qubits and lack of fidelity of their operations. However, many problems require more complex computations than supported on such NISQ devices. In this section, the principle mechanisms of how to support such complex computations on NISQ devices is described.

## 6.1 The Main Idea

Some quantum algorithms consist of both, a pure quantum part and a pure classical part (e.g. Shor's algorithm for factorization): its quantum part is performed on a quantum computer, and its classical part is performed on a classical computer. I.e. such an algorithm inherently requires classical pre- or post-processing to achieve its goal, it won't succeed without its classical part. Such a split algorithm is referred to as a *hybrid quantum-classical algorithm*.

Instead of such algorithms that are inherently hybrid, algorithms might be designed to be hybrid from the outset to limit their processing on a quantum computer. For example, this is enforced by today's near-term quantum devices that restrict the amount of processing that can be performed with sufficient precision. Such an algorithm requires additional classical processing to compensate for the restricted amount of work performed by the quantum computer: the algorithm is not inherently hybrid quantum-classical but it is so by design.

Two main problem domains are addressed by such kind of hybrid algorithms by design: problems that at their core can be solved based on the Raleigh-Ritz principle (Yuan et al. 2019), and problems that can be reduced to combinatorial optimization problems. Both kind of algorithms make use of a quantum part that consists of parameterized quantum circuits that prepare a state on a quantum computer and measure it, and a classical part that optimizes the corresponding parameters in iterations in dependence of the measured results (see Figure 16). The parameterized unitary operator $U(p_1, \ldots, p_k)$ in Figure 16 is called an "ansatz": its goal is to prepare states that deliver measurement results m that are good candidates for arguments optimizing a given function F. Thus, the value F(m) is dependent on the parameters $p_1, \ldots, p_k$. By varying these parameters based on classical optimization algorithms a corresponding m may be found (Shaydulin et al. 2019). Finding an appropriate ansatz is hard. Note, that a cloud environment is particularly suited to perform hybrid quantum-classical algorithms (Leymann et al. 2020, Karalekas et al. 2020).

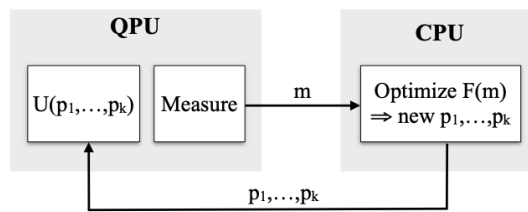

**Fig. 16**. Principle structure of variational hybrid quantum-classical algorithms.

Varying the parameters iteratively gives rise to the name *variational* hybrid quantum-classical algorithm (or variational algorithms for short) for this kind of approach. The application areas of variational algorithms span a broad spectrum including factoring of numbers (Anschuetz et al. 2018), solving linear equations (Bravo-Prieto et al. 2020), training autoencoders (Romero et al. 2017), or solving non-linear partial differential equations (Lubasch et al. 2020). A general approach towards a theory of variational hybrid quantum-classical algorithm is proposed in (McClean et al. 2016).

## 6.2 Maximum Cut: A Combinatorial Optimization Problem

The maximum cut problem is to partition the node set of a graph in two sets such that the number of edges between nodes of the different sets is maximal. In Figure 17, part 1, a graph with four edges {A,B,C,D} is shown where each node is connected to each other node. Part 2 and part 3 of the figure depict two different maximum cuts: the members of one node set are colored black while the members of the other node set are left white. Dashed edges connect nodes of the same set, i.e. they do not contribute to the cut, while solid lines connect nodes of the two different set, i.e. they contribute to the cut. The number of edges of each cut is four.

Part 4 of the figure adds a weight to each edge, e.g. the weight of an edge indicates the distance between the nodes connected by the edge; such a distance, for example, might be determined by means of a distance measure (see section 4.2). The weighted maximum cut problem strives towards finding a partition of the node set such that the sum of the weights of the edges between nodes of different sets is maximal. Note, that by setting each weight to "1", the maximum cut problem is seen to be a special case of the weighted maximum cut problem. Part (4) shows a weighted maximum cut: the node set is partitioned into {A,B,C} and {D}, and the sum of the weights of the corresponding edges is 20. If the weights are distances, the weighted maximum cut is nicely interpreted as determining clusters: the two node sets consist of nodes that are close to each other within each set, but nodes of different sets are far apart. For example, D has distance 6, 9, 5 to A, B, C, respectively, but A, B, C are quite close to each other (distances 1, 1, 2). Thus, an algorithm for determining a weighted maximum cut can be used for clustering.

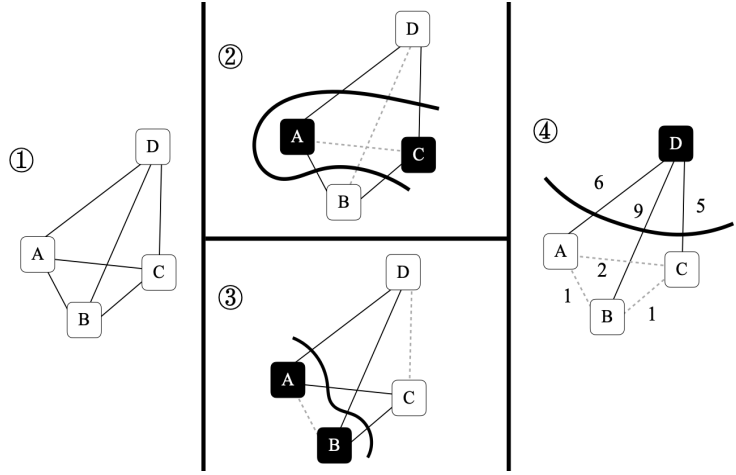

**Fig. 17.** Maximum cut of a graph: examples.

Let G=(N,E) be a graph with n nodes (i.e. card(N) = n), and let S, T be a cut of N (i.e. S $\cap$ T = $\varnothing$, S $\cup$ T = N, S $\neq$ $\varnothing$, and T $\neq$ $\varnothing$). The function $z: N \rightarrow \{-1, +1\}$ indicates membership to S or T, i.e. $z(u) = +1 \Leftrightarrow u \in S$ and $z(u) = -1 \Leftrightarrow u \in T$ (elements of S are labelled +1, elements of T are labelled $-1$ - or "black" and "white" in the figure). For brevity, $z_u := z(u)$.

For i, j $\in$ S, it is $z_i$=$z_j$=+1, thus $z_i \cdot z_j = 1$; and for i, j $\in$ T, it is $z_i$=$z_j$=$-1$, thus $z_i \cdot z_j = 1$. Consequently, it is i,j $\in$S $\vee$ i,j $\in$T $\Leftrightarrow$ $z_i \cdot z_j$=1, i.e. (i,j) is not a cut edge. Similarly, (i$\in$S $\wedge$ j$\in$T) $\vee$ (j$\in$S $\wedge$ i$\in$T) $\Leftrightarrow$ $z_i z_j$=$-1$, i.e. (i,j) is a cut edge. Reformulation results in:

- (i,j) $\in$ E is not a cut edge $\Leftrightarrow$ $1 - z_i z_j = 0$ $\Leftrightarrow$ ½·$(1 - z_i z_j) = 0$
- (i,j) $\in$ E is a cut edge $\Leftrightarrow$ $1 - z_i z_j = 2$ $\Leftrightarrow$ ½·$(1 - z_i z_j) = 1$

Thus, a maximum cut maximizes the following cost function:

$$C(z) = \frac{1}{2} \sum_{(i,j) \in E} \left(1 - z_i z_j\right)$$

This is because each cut edge contributes a "1" to the sum, and a non-cut edge contributes a "0": the more cut edges exist, the higher is C(z). Such kind of a formula, i.e. a formula with parameters with values from $\{-1, +1\}$ is called an *Ising formula*, which is important for solving quantum problems on so-called quantum annealers.

In this chapter it is assumed that algorithms are formulated in the gate model. For this purpose, parameters from {0,1} are better suited, i.e. the above cost formula has been transformed. To achieve this, a function $x: N \to \{0,1\}$ indicates membership to S or T, i.e. $x(u) = 1 \Leftrightarrow u \in S$ and $x(u) = 0 \Leftrightarrow u \in T$. For brevity, $x_u :=$ x(u). This implies:

- (i,j) ∈ E is not a cut edge $\Leftrightarrow$ i,j∈S ∨ i,j ∈T $\Leftrightarrow$ $x_i \cdot (1 - x_j) + x_j \cdot (1 - x_i) = 0$

- (i,j) ∈ E is a cut edge $\Leftrightarrow$

  (i∈S ∧ j∈T) ∨ (j∈S ∧ i∈T) $\Leftrightarrow$ $x_i \cdot (1 - x_j) + x_j \cdot (1 - x_i) = 1$

The cost formula with binary parameters for a maximum cut is then:

$$C(x) = \frac{1}{2} \sum_{(i,j) \in E} \left( x_i \cdot (1 - x_j) + x_j \cdot (1 - x_i) \right)$$

The term $x_i \cdot (1 - x_j) + x_j \cdot (1 - x_i)$ is a map $\{0,1\}^n \to \{0,1\}$, which is also called a *clause*. A combinatorial optimization problem is to determine a $z \in \{0,1\}^n$ (i.e. binary parameters) that maximizes (or minimizes) a cost function

$$C(z) = \sum_{\alpha=1}^{m} C_\alpha$$

where each $C_i : \{0,1\}^n \to \{0,1\}$ (1≤i≤m) is a clause with n binary parameters. This reveals that the maximum cut problem is a (binary) combinatorial optimization problem. Many other practically relevant problems like the traveling salesman problem are such problems. Next, it is shown how such problems can be solved with quantum computers.

## 6.3 QAOA

The concept of the quantum approximate optimization algorithm (QAOA) has been introduced by (Farhi et al. 2014). It is used to approximately solve combinatorial optimization problems on near-term quantum devices, and has been first applied to the maximum cut problem in (Farhi et al. 2014). (Crooks 2018) provided a QAOA-based implementation of a maximum cut algorithm on a near-term quantum computer of Rigetti and showed that this implementation is faster than the well-known classical Goemans-Williamson algorithm. The results can even be improved following the warm starting idea for QAOA (Egger et al. 2020).

The basic idea is as follows: The cost function C of a combinatorial optimization problem induces a map $C: \mathbb{H}^n \to \mathbb{H}^n$ by interpreting $z \in \{0,1\}^n$ as the computational basis vector $|z\rangle \in \mathbb{H}^n$ and defining

$$C|z\rangle := \sum_{\alpha=1}^{m} C_\alpha(z) \cdot |z\rangle = f(z) \cdot |z\rangle$$

Thus, the matrix of C in the computational basis is a diagonal matrix, i.e. the map C is Hermitian and can be used to measure the states of a quantum computer (C is an observable). Let $z' \in \{0,1\}^n$ such that $f(z') = \max\{f(z)\} = C_{max}$. The expectation value $\langle C \rangle_\psi$ of measuring C in any state $|\psi\rangle = \Sigma x_z |z\rangle$ is $\langle C \rangle_\psi = \langle \psi | C\psi \rangle = \langle \Sigma x_z |z\rangle | \Sigma x_z f(z) |z\rangle \rangle = \Sigma |x_z|^2 f(z) \quad\quad \leq \Sigma |x_z|^2 f(z') = f(z') \Sigma |x_z|^2 = f(z') = C_{max}$, i.e. the expectation value is less than or equal $C_{max}$. If $|\psi\rangle$ is close to $|z'\rangle$, $\langle C \rangle_\psi$ will be close to $C_{max}$. This means that with

$$|\psi\rangle = \sum_{z \in \{0,1\}^n} x_z |z\rangle = \sum_{z \neq z'} x_z |z\rangle + x_{z'} |z'\rangle$$

it must be achieved that $|x_{z'}|^2$ gets close to 1 (so-called amplitude amplification), i.e. a measurement will result with high probability in $z'$. This is realized by a unitary transformation (the "ansatz" - see section 6.1) the construction of which involves the cost function and rotations that emphasize components $x_z$ of the state $\psi$ the more $z$ contributes to the cost; these angles are the parameters of the ansatz that are classically optimized. For all the details see (Farhi et al. 2014).

## 6.4 Computing Maximum Cuts Via Eigenvalues

Let G=(N,E) be a graph with n nodes (i.e. card(N) = n), and let S, T be a partition of N. Such a partition can be described by a bit vector $x \in \{0,1\}^n$ as follows: $x_i = 1 \Leftrightarrow i \in S$ and $x_i = 0 \Leftrightarrow i \in T$. There are $2^n$ different bit vectors, i.e. partitions. For each partition the number $w_d$ of edges between nodes of different sets of the partition as well as the number $w_s$ of edges between nodes of the same set of the partition is determined. Next, $x \in \{0,1\}^n$ is considered as the binary representation of a natural number $x \in \mathbb{N}$, and $w_x := \frac{1}{2}(w_s - w_d)$ is associated with this number x. Finally, the matrix $M \in \mathbb{C}^{2^n \times 2^n}$ is defined where $m_{ij} = 0 \Leftrightarrow i \neq j$ and $m_{ii} = w_i$, i.e. the i-th row of M has as diagonal element the number $w_i$ corresponding to the partition $i \in \{0,1\}^n$, and all other elements of the row are zero.

M is a diagonal matrix and, thus, Hermitian, i.e. all eigenvalues are real numbers. Each vector of the computational basis is an eigenvector of M. It can be proven, that each vector of the computational basis that is an eigenvector of the lowest eigenvalue defines a maximum cut. Thus, determining the lowest eigenvalue of the matrix M and one of its eigenvectors means to determine a maximum cut of the corresponding graph. Note, that an analogue construction can be made for arbitrary weighted graphs. Using eigenvalues to solve the maximum cut problem is based on (Poljak and Rendl 1995).

## 6.5 VQE

A variational hybrid algorithm (called variational quantum eigensolver VQE) to determine the lowest eigenvalue of an operator has been first developed by (Peruzzo et al. 2014). This work has been extended towards a variational hybrid algorithm to approximate all eigenvalues of an operator (Higgott et al. 2019). As a consequence of section 6.4, the maximum cut problem can be solved by using VQE. In addition, dimension reduction based on principal component analysis (PCA) requires the computation of all eigenvalues of a certain matrix, i.e. this algorithm is key for our work.

The basic idea of VQE is as follows: according to the Raleigh-Ritz principle (Yuan et al. 2019), the lowest eigenvalue $\lambda_{min}$ of a Hermitian operator O satisfies the following inequation:

$$\lambda_{min} \leq \frac{\langle \psi|O|\psi \rangle}{\|\langle \psi, \psi \rangle\|} \text{ for any } |\psi\rangle \neq 0.$$

With $\langle \psi|O|\psi \rangle = \langle O \rangle_{|\psi\rangle}$ for each state $|\psi\rangle$ (i.e. $\| \psi \| = 1$), the expectation value of O provides an upper bound of the lowest eigenvalue of O: $\lambda_{min} \leq \langle O \rangle_{|\psi\rangle}$. Thus, if a state $|\psi\rangle$ can be found that minimizes the expectation value of O, $\langle O \rangle_{|\psi\rangle}$ is close to the lowest eigenvalue of O. (Note, that $\langle O \rangle_{|\psi'\rangle} = \lambda_{min}$ holds for an eigenvector $|\psi'\rangle$ of $\lambda_{min}$).

In order find such a $|\hat{\psi}\rangle$, series of states $|\psi(p_1, \ldots, p_k)\rangle$ are iteratively constructed that depend on parameters $p_1, \ldots, p_k$; in Figure 16 the unitary operator $U(p_1, \ldots, p_k)$ prepares this state. The expectation value $\langle O \rangle_{|\psi(p_1, \ldots, p_k)\rangle}$ is measured for each parameterized state, and the measurement result is subject to a classical optimization algorithm that determines new parameters $p_1, \ldots, p_k$ that reduce the expectation value further; in Figure 16 the function F(m) is this expectation value. These iterations approximate $|\psi'\rangle$, i.e. the final set of parameters determine $\langle O \rangle_{|\psi(p_1, \ldots, p_k)\rangle} \approx \lambda_{min}$ and $|\psi(p_1, \ldots, p_k)\rangle$ is an approximation of the corresponding eigenvector.

An efficient measurement of the expectation values is done as follows: Each Hermitian operator O can be written as a linear combination of "simpler" Hermitian operators $O_i$, i.e. $O = \Sigma x_i O_i$. Here, "simpler" means that $O_i$ is a combination of Pauli operators. The expectation value $\langle O_i \rangle_v$ of such a combination can be efficiently measured in a quantum computer (Hamamura and Imamichi 2019), and it is $\langle O \rangle_v = \Sigma x_i \langle O_i \rangle_v$, i.e. the expectation value of O can be easily computed classically based on the measured $\langle O_i \rangle_v$ - i.e. in Figure 16 the function F computes $\Sigma x_i O_i$. For all the details see (Peruzzo et al. 2014).

# 7 QHAna: A Quantum Humanities Analysis Tool

As described in sections 5 and 6 there are several machine learning techniques for which first implementations are available on quantum computers. Some can be used to just gain initial experiences, others already show improvements in terms of speedup or precision. Bringing together the aspects and concepts outlined in the above sections, QHAna (QHAna 2021) was developed. QHAna is our quantum humanities data analysis tool that aims for several goals: (i) There is the content-related goal of supporting the identification of patterns by data analysis. A feasibility study is provided by the use case (section 3) that focuses on analyzing costume data to improve the understanding of vestimentary communication in films. (ii) By performing this analysis by classical and hybrid quantum-classical algorithms a comparison of both methods is supported, allowing to assess the benefits of using quantum algorithms. (iii) This comparison supports the goal of improving the understanding of potential benefits quantum computing may provide for digital humanities research (see section 2.2). (iv) Additionally, QHAna allows the integration of heterogeneous tools from different quantum computing vendors in a single analysis pipeline. (v) Thus, one of the overarching goals of QHAna is to provide easy access to people without quantum computing background to gain first application knowledge in the domain of quantum humanities. In the following QHAna will be introduced by this core objectives and functions.

## 7.1 Support the Identification of (Costume) Patterns

The primary goal of QHAna is to support analyzing data when aiming at the identification of patterns based on the data analysis pipeline defined in section 4.1. By providing easy access to the analysis methods and techniques outlined above, QHAna is designed to e.g., improve the understanding of vestimentary communication in films by supporting the identification of costume patterns in the MUSE dataset (see section 3.3). Note that QHAna is data independent and therefore, not limited to the data of our use case. Data from other application areas like from the MUSE4Music project (Barzen et al. 2016) is planned to be analyzed.

Based on initial question like (i) "which base elements are used to communicate specific stereotypes?", (ii) "do certain age impression and attributes like color or condition communicate certain character traits within one genre?", (iii) "and if this is true, how to group new costumes to the suitable category when being captured?" several analysis steps need to be performed. The upper part of Figure 18 gives an impression of the main menu that guides through the pipeline for analyzing data. To approach an answer to these sample questions, for example, question (ii) is addressed by clustering algorithms, focusing on finding those costumes that achieve the same effect, while the question (iii) is addressed by classification algorithms. For both, the first step is about preparing the data depending on the requirements of the algorithms. As this is the basis for all the other following steps, it will be described in more detail how QHAna supports this step.

As Figure 18 depicts, the tab "Data Preparation" allows to prepare the data depending on the requirements of specific analysis algorithms. As described in section 4.2, several algorithms require numerical data and the MUSE data is mostly categorical data. Therefore, the subtabs "Distance Space Calculation" and "One-Hot Encoding" as shown in Figure 18 provide different options to transform categorical data into numerical ones. For our example we chose the "Distance Space Calculations", that supports the approach described in section 4.2: Here, a dropdown menu allows to choose the domain of the distance space. In our MUSE use case, the "Costume" data set is chosen. Figure 18 depicts how to specify a view for analyzing a defined subset of the attributes of the MUSE data set. For this purpose, the attributes of interest of our example question are chosen, e.g., "Dominant Color", "Dominant Condition", "Dominant Character Trait", "Dominant Age Impression", and "Genre" are selected. Furthermore, the user can define which element comparer (that allow to compare two elements within an attribute category) and attribute comparer (that allow to compare sets of elements within an attribute), or which aggregator type (that define how data points are aggregated) and transformer type (that describe the function used to transform similarity measures into distance measures), are most suitable for the use case, as well as how empty attributes are to be treated (e.g. if a costume has no value for the attribute category "Color" the missing element should be "ignored"). The table at the top provides an overview of all selections made.

The tap "Documentation" (right of "Data Preparation") allows to retrieve general information about the currently processed data and its structure (e.g. attribute lists, taxonomies), explanations about the components already implemented in QHAna (e.g. different cluster or classification algorithms), references to software dependencies as well as all relevant related papers used to implement the tool.

**Fig. 18**. Screenshot of QHAna "Data Preparation"

After computing the distance space, QHAna supports calculating corresponding distance matrices (see section 4.2). Here, the user can choose how many costumes should be analyzed and whether these costumes should be selected randomly from the whole data set or custom specific. Figure 19 gives an example of such a distance matrix of 25 costumes. Note that the visualization of the distance may become very complex with an increasing number of input data and the resulting visualization may become difficult to comprehend.

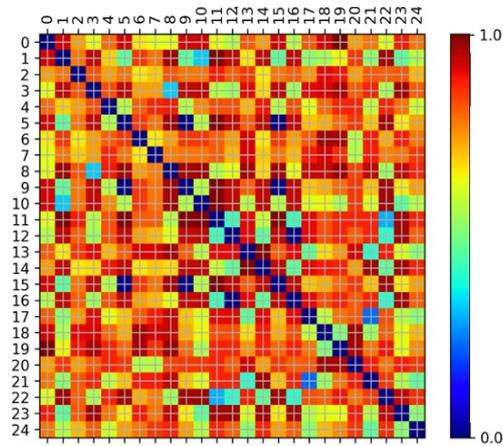

**Fig. 19.** Screenshot of a distance matrix.

Distances between pairs of data elements (e.g. costumes) are represented by a color code: The distance value of two costumes becomes a coefficient of the distance matrix and this value is turned into a color: dark blue corresponds to no distance at all, while dark red indicates maximum distance. Each costume is represented by a number (0-24 in Figure 19). As the distance matrix depicts, some of the costumes are very close to each other based on the selected attributes and others are very far away (i.e. they are not really similar): For example, costume 5 is very close to costume 9 and costume 15, while it has hardly any similarity with (i.e. it is far away from) costume 11 and 22. For a more detailed comparison of these costumes the "Entity Table" subtab of the "Overview" tab (see Figure 20) allows to compare the costumes of interest in detail by their attributes chosen. Figure 20 depicts costumes 5, 9, 11, 15 and 22 stressing that 5, 9 and 15 have the same attribute values, while costume 11 and 15 differ in their attribute values. As most costumes have more than just 5 attributes, changing the input parameters in the costume distance space definition may have a deep impact on the similarity measures of these costumes. Also, experts sometimes need to verify the concrete costumes manually. For this purpose, the entity table provides for each costume links to the MUSE-repository (the film, the character, the concrete costume) allowing further investigation of the costumes with all the detailed information, in particular it provides the visual representations of the costumes from the MUSE-repository.

After the data is prepared the next step allows to – if needed – perform "Feature Engineering" to reduce the number of relevant attributes. The tab "Feature Engineering" provides several techniques for dimension reduction and feature extraction such as autoencoders, embeddings, or PCA. This provides the basis to use clustering techniques that allow to group those costumes together that have the same effect, e.g. communicating a Sheriff by certain base elements like a Sheriff star, Cowboy boots and so on (Barzen 2018). As there are new costumes captured on a daily basis and they need to be classified in terms of being mapped to the costume pattern they contribute to, running classification algorithms is very promising for the MUSE use case. All those steps supported by QHAna aim at improving the understanding of vestimentary communication in films. How feature engineering, clustering, and classification are supported is unfolded in what follows.

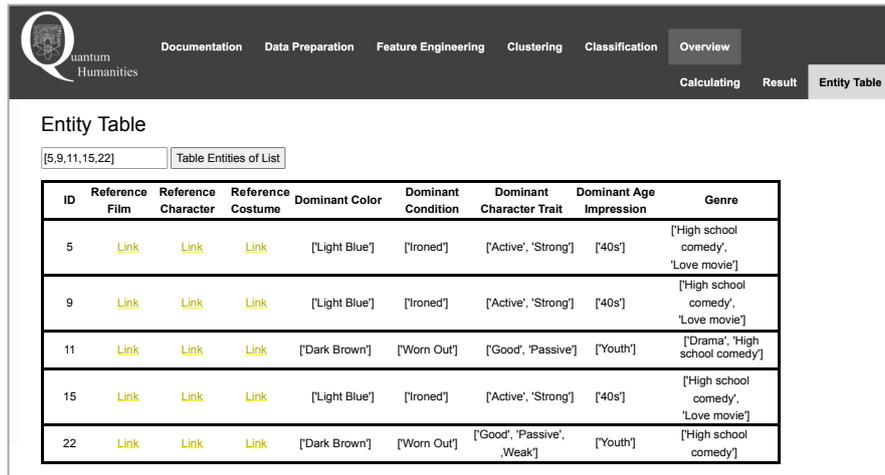

**Fig. 20**. Screenshot of "Entity Table".

## 7.2 Comparing Classical and Quantum Machine Learning Algorithms

To get the optimal results when approaching an answer to a stated question the different analysis steps are performed based on different algorithms, often in parallel by different classical implementations as well as by different quantum-classical implementations. This allows comparing the different results of classicals machine learning algorithms amongst each other, comparing different results of quantum machine learning algorithms amongst each other, and comparing results from classicals machine learning algorithms with results from quantum machine learning algorithms. This enables finally selecting the most suitable algorithm for the problem at hand as well as improving the application of an algorithms when comparing the results and optimizing iteratively the hyperparameters. Please note, that due to the limitations of today's quantum computers only small sets of data can be processed when using quantum machine learning algorithms.

To outline how such a comparison is enabled by QHAna, two examples are provided, one from "Feature Engineering" where a different technique is used to get so a distant matrix and one from "Clustering" where the comparison of different implementations of maximum cut algorithms (see section 6.2-6.5) is outlined.

**Classical and Quantum-based Approaches to Distance Matrices**

As described in section 5.4 and 5.5, autoencoders can be used for dimension reduction. Therefore, four implementations of autoencoders have been used to reduce the dimension of MUSE data. We realized two autoencoders classically (based on (PyTorch 2021) and (TensorFlow 2021)) and two autoencoders in a hybrid manner: in the latter case, a classical autoencoder is first used to reduce all input data to a three-dimensional feature space, while in the succeeding step (i) a quantum inspired autoencoders (using (TensorFlow Quantum 2021)) and (ii) a quantum autoencoders (using Qiskit, see Figure 15 for an impression of the circuit) are used to further reduce the dimension to two (for more details see (Barzen et al. 2021). Please note that those autoencoders are currently being integrated into QHAna). By pairwise measuring the fidelity of the quantum states and applying the Fubini-Study metric (Biamonte 2020) to the measured data, the distances between the resulting quantum states can be computed.

The left side of Figure 21 gives an example of the distance values of 10 costumes (with the same 5 attributes selected as in Figure 18) as result of a hybrid quantum inspired autoencoder. As before, 0 (dark blue) corresponds to the smallest distance and 1 (bright red) to the largest possible distance (note, that based on the actual data, the largest distance is about 0.30). What can be seen is that the costumes 0 to 4 are highly similar to each other but rather different from costumes 5 to 9, and that the costumes 5 to 9 are again highly similar to each other. The right side of Figure 21 shows the distances of the same 10 costumes determined by using the Wu and Palmer similarity measure (see 4.2 and 7.1). What can be seen is that quite similar results have been achieved by using these two different techniques: The statistical approach (autoencoder) on the one hand and the approach using taxonomies (Wu-Palmer similarities) on the other hand, identified the same costumes as being close to each other. This hints to a potential costume pattern. Also, it provides first insights on comparing a classical and a quantum-based approach to compute distance values.

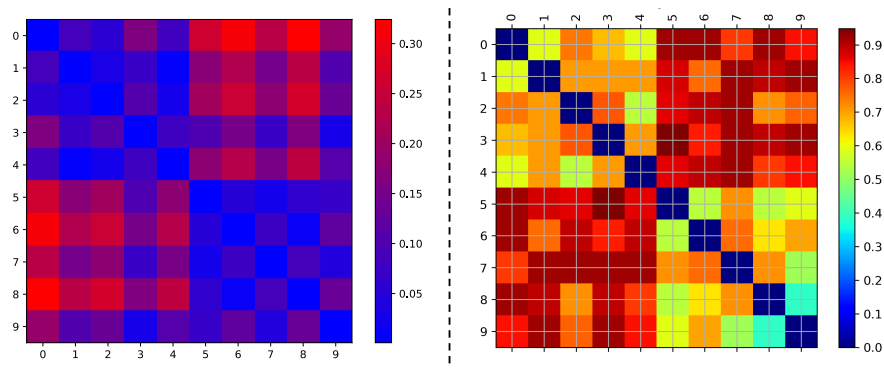

**Fig. 21.** Distance matrix as result of a hybrid autoencoder (left side) and computed via the Wu and Palmer similarity (right side).

## Classical and Quantum-based Approaches to Clustering

Implementations of both, classical clustering algorithms and first quantum-classical clustering algorithms are available under the tab "Clustering" in QHAna. Currently, this includes an implementation of the OPTICS cluster algorithm (Ankerst et al. 1999), four different implementations of the maximum cut (a.k.a. maxcut) algorithm (see section 6.2) and four different implementations of the k-means algorithm (Arthur and Vassilvitskii 2007, Khan et al. 2019). The implementations of the maximum cut range from a classical naive implementation ("classicNaiveMax-Cut"), to two approximative implementations ("sdpMaxCut" based on semidefinite programming (Boumal 2015) and "bmMaxCut" using a Bureir-Monteiro solver (Boumal et al. 2016)), to a quantum-classical implementation ("qaoaMaxCut") that is based on QAOA (see section 6.3) and implemented in Qiskit (see Max Cut 2021).

Figure 22 allows a comparison of the results of using different maximum cut implementations. Input are the same 10 costumes as in Figure 21, using the default values of the hyperparameters per algorithm provided by QHAna. Diagram 1 of the figure shows the costumes and their distances (upper left corner) in an MDS-based embedding. The subtab "Embedding" of "Feature Engineering" supports to use MDS as one implemented approach to map the distance measures of the chosen costumes with their chosen attributes to a metrical feature space. As a result of the embedding via MDS the data is mapped into $\mathbb{R}^n$ in such a way that the original distances of the data elements are nearly the distances in the feature space (see section 4.2). The other three diagrams present clustering results achieved by different maximum cut implementations namely "qaoaMaxCut" (diagram 2), "sdpMaxCut" (diagram 3), and "bmMaxCut" (diagram 4). It can be seen that the two approximative implementations (diagram 3 and 4) have identical results: The first cluster (red circles) contains the costumes 0-4, while the second cluster (blue circles) contains the costumes 5-9. The result of the quantum-classical implementation (part 2) is rather different: cluster one (red circles) contains the costumes 3, 4, 7, and 9 while the second cluster (blue circles) contains the costumes 0, 1, 2, 5, 6, and 8. In providing easy access to several implementation the comparison of such results is supported by QHAna. In addition, the results can be manually verified using the entity table of QHAna (see Figure 20) as outlined in section 7.2. The entity table allows the evaluation of the attributes of the costumes that are part of the clusters and provides all the details of the costumes by linking them to the MUSE-repository. This can be used to improve the understanding of the benefits quantum computing can have, e.g. for quantum humanities research.

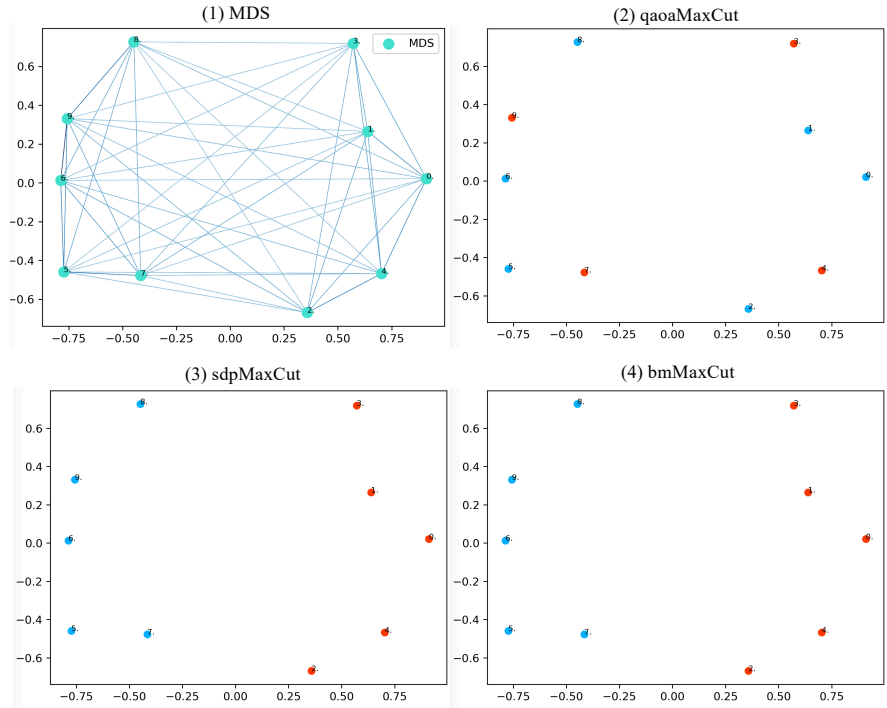

**Fig. 22.** Values of 10 costumes embedded via MDS (1) and their clustering results of different maximum cut implementations (2-4).

### 7.3 Improve the Understanding of the Benefits of Quantum Computing

In a fist attempt when taking a closer look at the comparison of Figure 22 one might decide for the classical implemented clustering for further analysis as it seems more precise. But QHAna also allows to improve the results by adjusting the hyperparameters of the algorithms. For the results of Figure 22 this would include to potentially improve the results of the quantum-classical implementation of the algorithm as shown in Figure 22 to equal the results of the classical implementation. Thus, QHAna lists the selectable parameters (a.k.a. hyperparameters) specific to the chosen algorithms, together with descriptions that support selecting the right parameters. Figure 23 gives an impression of the hyperparameters to be selected for the quantum-classical implementation of a maximum cut algorithm including the user token required to run the algorithm on the IBM Q systems.

**Fig. 23.** Screenshot of "Clustering".

Figure 24 outlines the influence of the maximum number of iterations performed by the algorithm, while all other parameters keep the default settings of QHAna. As can be seen, increasing the number of iterations seems to improve the results: Diagram 1 is performed with only one iteration, while the number of maximum iterations is increased by 50 each time (diagram 2: 50 trials, diagram 3: 100 trials, and diagram 4: 150 trials). Thus, the results approach those achieved by the classical maximum cut implementations up to the identification of the same clusters when performed with a maximum of 150 trials (diagram 4).

By enabling easy comparison of different techniques and implementations of classical and quantum-classical algorithms during the data analysis process, QHAna contributes to a better understanding of the potential benefits (section 2.2) that quantum computing can provide for quantum humanities research. This includes as described above identifying optimal hyperparameters, e.g., approximating the results of the quantum-classical implementation and the classical ones, which then can be applied to larger use cases. But it also includes identifying already available advantages of applying quantum machine learning. An example of these potential advantages of quantum machine learning, is the precision that the quantum kernel estimation (QKE) method for SVM (Havlicek et al. 2018) can achieve.

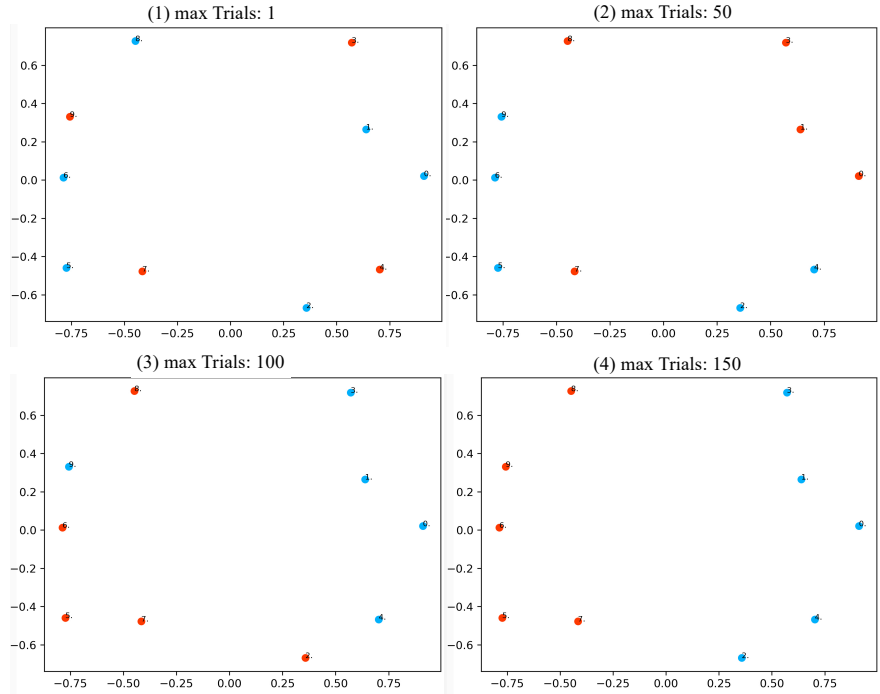

**Fig. 24**. Clustering results of the quantum-classical maximum cut implementations when changing the number of maximum iterations performed.

Currently the tab "Classification" of QHAna provides three implementation of SVM (see section 5.3) methods ("classicSklearnSVM" an implementation of classical support vector machines, "qkeQiskitSVM" an implementation using the quantum kernel estimation (QKE) variant of the SVM, and an implementation of an variational quantum SVM (Havlicek et al. 2018)) and two classifiers based on neural networks (an implementation based on a classical neural network, an implementation based on a hybrid quantum-classical neural network). Figure 25 gives an example of the results of the classically implemented SVM (diagram 1) and of the QKE based SVM (diagram 2). Input for this example are 30 costumes and their distance values based on the attributes stereotype and role relevance embedded by MDS, and classified as "positive" or "negative" based on another training set for the classifier. As can be seen in Figure 25 these classes are not linear separable. Thus, finding the right classifier is a challenging task. Figure 25 depicts that (by using the default parameters QHAna provides) the result of the QKE variant (diagram 2) is much more precise (accuracy of 94% versus 89%) than the result of the classical variant (diagram 1).

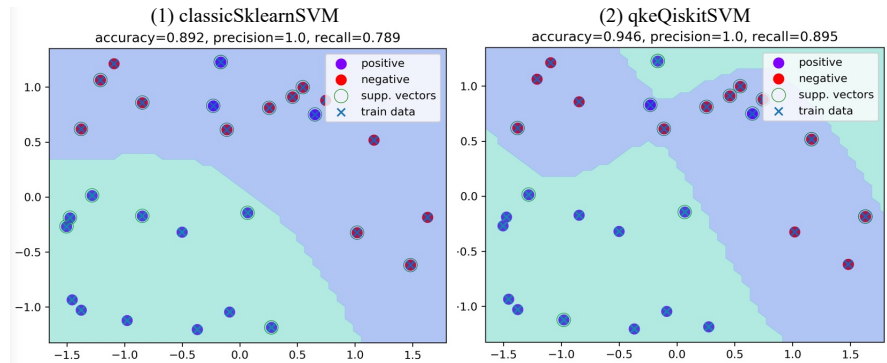

**Fig. 25**. Classification results of the classically implemented SVM (left) and of the QKE based implemented SVM (right).

### 7.4 Integration of Heterogeneous Tools

The prototype of QHAna has a modular structure. Thus, it can easily be extended by further algorithm implementations. Various implementations are already available (see section 7.1-7.3). Because QHAna can also serve REST APIs the effective integration of components accessible via such APIs is straightforward; this way, further implementations of techniques described in the theoretical sections of this contribution into QHAna is achieved: This includes – but is not restricted to – extending the feature engineering step by quantum principal component analysis, for example, or add quantum restricted Boltzmann machines, using more clustering algorithms and integrating more classification algorithms. Also, algorithms used in other domains (like HHL, VQE and so on) can be integrated into QHAna.

Another significant advantage of QHAna is, that it provides comfortable access to different backends to run the algorithms on. Figure 23 shows, for example, that there are different simulators and IBM quantum computing backends integrated and selectable already. Currently we are integrating more backends like PennyLane (PennyLane 2021) and TensorFlow.

### 7.5 Provide Easy Access to Quantum Computing

Working with quantum computers still requires advanced knowledge in physics and mathematics and is therefore a great challenge for research in the field of digital humanities. As there are promising potentials in using the quantum computer in this area (see section 2.2), it is of interest to provide a simple, straightforward access to this new technology without having to deal with all the mathematical, physical, and algorithmic details. This is what QHAna aims for. As Figure 23 depicts, even very complex quantum-classical algorithms are provided in an abstract manner allowing to adjust their hyperparameters, but default values of these parameters are set. Thus, these algorithms and their benefits can be applied by someone not familiar with qubits, QPU connectivity, quantum circuits, and so on. In sections 5 and 6 we provided the core concepts to enable an understanding of the underlying ideas of the (classical and quantum) algorithms used in QHAna, but no details about classical algorithms nor quantum algorithms are given. These algorithms and their implementations evolve rapidly and potentially change quite frequently. Their details are not the focus of QHAna. Instead, providing access to a complex new technology to learn about and participate in the advantages given for further research is at the heart of QHAna.

## 8 Conclusion and Outlook

As outlined in this contribution there are a lot of potentials in applying quantum computers to research done in the humanities but there is also a long way to go to fully benefit from them. Thus, the benefits and challenges of quantum computers were stressed and – based on our use case MUSE – first application knowledge for further quantum humanities is provided.

As the use case is focused on machine learning, a data analysis pipeline for machine learning tasks was introduced (that is especially useful for detecting patterns in a domain) and core concepts that are promising to be applied in the fields of the humanities were introduced and discussed. Therefore, artificial neural networks were described by providing a mathematical definition of neurons, neural networks, and perceptrons. Also, their use for restricted Boltzmann machines and autoencoders as well as first realizations of these on quantum computers were described. Quantum algorithms are often hybrid; therefore, the main idea of variational hybrid quantum-classical algorithms was sketched as well as an application for clustering based on hybrid techniques like quantum approximate optimization algorithm (QAOA) and variational quantum eigensolver (VQE) were discussed.

To provide a straightforward access to the described techniques an analysis tool called QHAna has been introduced. The presented prototype has a modular architecture and implementation; thus, it can be extended easily. We plan to realize an import and export functionality such that - after key steps of the supported pipeline - the intermediate results can be exported, and results from other steps can be imported. Especially, this will allow to consider results from external algorithms (i.e. algorithms not available via QHAna) in the pipeline as well as to pass results from QHAna for further processing to external algorithms. We also plan to integrate workflow features such that users may define their own sequence of invoking algorithm implementations (those provided by QHAna as well as external ones) to support custom pipelines of data analysis in a multitude of domains: this will significantly improve the flexibility of QHAna.

As quantum computers are continuously improving and the number of qubits is expected to increase constantly and significantly (Gambetta 2020), it is of great importance to develop application knowledge in the domain of quantum computing at an early stage. Therefore, we envision to generalize the application knowledge that can be extracted from our approach to *quantum humanities patterns* to make it reusable for different use cases in other application domains of the humanities. Providing solution knowledge on such a promising and innovative topic as quantum computing is already relevant in itself. First samples for pattern languages in this domain can be seen (Leymann 2019, Weigold et al. 2020, Weigold et al. 2021). It is of particular relevance to provide application knowledge in a field where the mathematical and physical basics required to use this technology cannot be taken for granted – but fields, which are essential for the critical reflection of digital methods, such as the digital humanities. Especially, if a quantum computer is not only to be used as a tool like for quantum machine learning but for thinking about totally new questions that have not yet been tackled at all, perhaps not even identified or considered, knowledge about the corresponding concepts, methods, and application potentials is key. To identify and explore these possible new application areas is an exciting task and needs to be started by building knowledge on how quantum computers can be applied to problems stated in the humanities.

## 9 Acknowledgement

I am very grateful to Frank Lemann for discussing several subjects of this chapter. Also, I would like to thank Felix Truger, Philipp Wundrack, Marcel Messer, Daniel Fink and Fabian Bühler for their valuable input and implementing several aspects of our use case.

This work was partially funded by the BMWi project PlanQK (01MK20005N) and the Terra Incognita project Quantum Humanities funded by the University of Stuttgart.

(All links have been last followed March 11, 2021)